\documentclass[5p]{elsarticle}

\journal{Journal of Robotics and Autonomous Systems}
\usepackage[dvipsnames]{xcolor}  
\usepackage[utf8]{inputenc}
\usepackage{graphics} 
\usepackage{epsfig} 
\usepackage{textcomp}
\usepackage{hhline}
\usepackage{mathptmx} 
\usepackage{times} 
\usepackage{amsmath} 
\usepackage{amssymb}  
\usepackage{bm}
\usepackage{tikz}
\usepackage{pgfplots}
\usepackage{multicol}
\usepackage{mathtools}
\usepackage{multirow}
\usepackage{float}
\usepackage{subfig}
\usepackage{wrapfig}
\usepackage{url}
\usepackage{nth}
\usepackage{caption}
\usepackage{algorithm}
\usepackage[noend]{algpseudocode}
\usepackage{balance}
\usepackage{dblfloatfix}
\makeatletter
\def\BState{\State\hskip-\ALG@thistlm}
\makeatother
\usepackage{float}
\usepackage[disable]{todonotes}

\usepackage{adjustbox}
\usepackage{array}
\newcolumntype{R}[2]{%
	>{\adjustbox{angle=#1,lap=\width-(#2)}\bgroup}%
	l%
	<{\egroup}%
}

\DeclareMathAlphabet{\mathcal}{OMS}{cmsy}{m}{n}

\newcommand{\TODO}[1]
{\todo[inline, author=todo, color=pink!50,bordercolor=pink!50,size=\scriptsize]{#1}}
\usepackage{blindtext}
\usepackage{siunitx}

\pgfplotsset{compat=newest}
\usepgfplotslibrary{units}
\newlength{\figwidth}          
\newlength{\figheight}         
\DeclareMathOperator*{\argmin}{argmin}

\newcommand{\eq}[1]{Equation~\eqref{#1}}
\newcommand{\fig}[1]{Figure~\ref{#1}}

\renewcommand{\sec}[1]{Section~\ref{#1}}
\newcommand{\tab}[1]{Table~\ref{#1}}
\newcommand{\alg}[1]{Algorithm~\ref{#1}}
\newcommand{\app}[1]{\ref{#1}}

\newcommand{\loss}[0]{\mathcal{L}}

\newcommand{\La}[0]{\loss_\text{act}}
\newcommand{\Lv}[0]{\loss_\text{vel}}
\newcommand{\Lp}[0]{\loss_\text{pos}}
\newcommand{\tta}[1]{\boldsymbol{\theta}_{#1}}
\newcommand{\tman}[1]{\boldsymbol{\theta}^\text{man}_{#1}}
\newcommand{\topt}[1]{\boldsymbol{\theta}^\text{opt}_{#1}}
\newcommand{\tlim}[1]{\boldsymbol{\theta}^\text{lim}_{#1}}
\newcommand{\plim}[1]{\mathbf{P}^\text{lim}_{#1}}
\newcommand{\pmax}[1]{\mathbf{p}^\text{max}_{#1}}
\newcommand{\pmin}[1]{\mathbf{p}^\text{min}_{#1}}
\newcommand{\pdes}[1]{\mathbf{p}^\text{des}_{#1}}
\newcommand{\R}[1]{\mathbb{R}^{#1}}

\begin{document}
	
	\begin{frontmatter}
		
		\title{Learning to Control Highly Accelerated Ballistic Movements on Muscular Robots}
		
		\author[mpiadd,tudaadd]{Dieter Büchler\corref{mycorrespondingauthor}}
		\author[fbadd]{Roberto Calandra}
		\author[mpiadd,tudaadd]{Jan Peters}
		
		
		%
		
		
		\address[mpiadd]{Max Planck Institute for Intelligent Systems, Max Planck Ring 4, 72076 Tübingen~(Germany)}
		\address[tudaadd]{Technische Universität Darmstadt, Hochschulstraße 10, 64289 Darmstadt~(Germany)}
		\address[fbadd]{Facebook AI Research, 1 Hacker Way, 94025 Menlo Park~(USA)}
		
		\cortext[mycorrespondingauthor]{Corresponding author}
		\begin{abstract}
High-speed and high-acceleration movements are inherently hard to control.
Applying learning to the control of such motions on anthropomorphic robot arms can improve the accuracy of the control but might damage the system.
The inherent exploration of learning approaches can lead to instabilities and the robot reaching joint limits at high speeds.
Having hardware that enables safe exploration of high-speed and high-acceleration movements is therefore desirable.
To address this issue, we propose to use robots actuated by Pneumatic Artificial Muscles (PAMs). 
In this paper, we present a four degrees of freedom~(DoFs) robot arm that reaches high joint angle accelerations of up to \SI{28000}{\degree\per\second\squared} while avoiding dangerous joint limits thanks to the antagonistic actuation and limits on the air pressure ranges. 
With this robot arm, we are able to tune control parameters using Bayesian optimization directly on the hardware without additional safety considerations. 
The achieved tracking performance on a fast trajectory exceeds previous results on comparable PAM-driven robots.  
We also show that our system can be controlled well on slow trajectories with PID controllers due to careful construction considerations such as minimal bending of cables, lightweight kinematics and minimal contact between PAMs and PAMs with the links.
Finally, we propose a novel technique to control the the co-contraction of antagonistic muscle pairs.
Experimental results illustrate that choosing the optimal co-contraction level is vital to reach better tracking performance. 
Through the use of PAM-driven robots and learning, we do a small step towards the future development of robots capable of more human-like motions.

		\end{abstract}
		
		
	\end{frontmatter}
	
	
	\section{Introduction}
	\label{sec:intro}
Controlling highly accelerated movements on anthropomorphic robot arms is an aspired ability.
High accelerations lead to high velocities over a small distance which enables fast reaction times.
Such motions can be observed in human arm trajectories, known as ballistic movements~\cite{zehr_ballistic_1994}.
However, producing ballistic movements on robots is challenging because they
1) are inherently hard to control,
2) potentially run the joints into their limits and hence break the system and
3) require hardware that is capable of generating high accelerations.
We use the term \textit{high-acceleration tasks} to refer to the set of such problems.



A promising way to approach problem 1) is to apply Machine Learning approaches - that inherently explore - to learn directly on the real hardware. 
In this manner, algorithms can automatically tune low-level controllers that, for instance, track a fast trajectory with lower control error than manually tuned controllers. 
Problem 2), however, currently rules this path out and is even more problematic once we generate higher accelerations~(Problem 3)).
Hence, enabling exploration in such fast domains by preventing potential damages from the hardware side, can help improve performance in high-acceleration tasks.
%
\begin{table*}[b]
	\caption{A collection of pneumatic based robotic arm-like systems, listed next to the number of DoFs in joint space and the fastest and most complex tracked trajectory if known.}
	\label{tab:pamarm}
	\centering
	\resizebox{\linewidth}{!}{%
		\begin{tabular}{c | c | c | c}
			Year& Publication  & \# DoF & Fastest and most complex trajectory tracked \\
			\hline
			2018 &Driess et al.~\cite{driess_learning_2018} & 2~(5 PAMs)  &  Reaching motions \\
			2016 &Das et al.~\cite{das_controlling_2016} & 2  &  Step signal to both DoFs\\
			2014 & Rezoug et al.~\cite{rezoug_experimental_2014}  & 7& Sinusoidal reference with $f=\SI{1}{\hertz}$ for one DoF\\
			2012 & Hartmann et al.~\cite{hartmann_real-time_2012}  & 7& Sinusoidal reference in task space~(x: $f=\SI{1}{\hertz}$, y: $f=\SI{2}{\hertz}$, z not tracked)\\
			2012 & Ikemoto et al.~\cite{ikemoto_direct_2012}  & 7~(17 PAMs) & Human taught reference~(similar to sinusoidal) periodic with $f\SI{\sim 0.33}{\hertz}$\\
			2009 & Ahn and Ahn~\cite{ahn_design_2009} & 2 & Triangular reference with $f=\SI{0.05}{\hertz}$\\
			2009 & Shin et al.~\cite{shin_design_2009}  & 1~(4 PAMs) & Sinusoidal reference with $f=\SI{6}{\hertz}$ for one DoF \\
			2009 & Van Damme et al.~\cite{van_damme_proxy-based_2009}  & 2~(4 PAMs) & Sinusoidal reference with $f=\SI{0.33}{\hertz}$ for both DoFs \\
			2007 & Festo Airic's arm~\cite{festo_airics_2007} & 7~(30 PAMs)  & N/A \\
			2006 & Thanh and Ahn~\cite{thanh_nonlinear_2006}  & 2 & Circular with $f=\SI{0.2}{\hertz}$ using both DoFs\\
			2005 & Hildebrandt et al.~\cite{hildebrandt_cascaded_2005} & 2 & Step and sinusoidal reference with $f=\SI{0.5}{\hertz}$ \\
			2005 & Tondu et al.~\cite{tondu_seven-degrees--freedom_2005}  & 7 & N/A\\
			2004 & Boblan et al.~\cite{boblan_human-like_2004} & 7 in arm & N/A\\
			2000 & Tondu and Lopez~\cite{tondu_modeling_2000} & 2  & Independent sinusoidal activation of each DoF with $f=\SI{0.1}{\hertz}$\\
			1998 & Caldwell et al.~\cite{caldwell_Pneumatic_1998}  & 7 & Response of shoulder joint to $\SI{90}{\degree}$ step reference\\
			1995 & Caldwell et al.~\cite{caldwell_control_1995}  & 7 & Response to a square wave reference input~($f=\SI{0.2}{\hertz}$, 1 DoF) \\   
		\end{tabular}
	}
\end{table*}
%

The human arm anatomy possesses many beneficial properties over current anthropomorphic motor-driven robotic arms for high-acceleration tasks. 
While motor-driven systems can generate high speeds, it is hard to produce high accelerations and keep the kinematics human arm sized at the same time.
Instead, muscles drive the human arm.
Skeletal muscles generate
\begin{figure}[H]
	\centering
	\includegraphics[width=\linewidth]{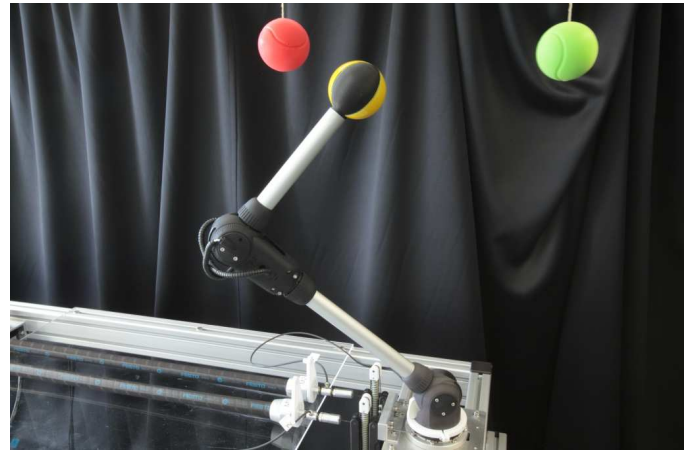}
	\caption{Igus Robolink lightweight arm with \SI{700}{\gram} of moving masses. 
		Eight powerful antagonistic PAMs move four DoFs where each joint contains two rotational DoFs. 
		Rather than recreating human anatomy, our system is designed to ease control to facilitate the learning of fast trajectory tracking control. 
		Experimental results show that our robot is precise at low speeds using a simple manually tuned PID controller while reaching high velocities of up to \SI{12}{\meter\per\second}~(\SI{200}{\meter\per \second\squared}) in task space and \SI{1500}{\degree\per\second}~(\SI{28000}{\degree\per\second\squared}) in joint space.
	}
	\label{fig:arm}
\end{figure} 
\noindent
high forces and are located as close as possible to the torso to keep moving masses to a minimum.
Concurrently, the human arm inhibits damage at collisions thanks to the built-in passive compliance which ensures deflection of the end-effector instead of breakage as a response to external forces.


Robotic arms actuated by antagonistic pneumatic artificial muscle~(PAM) pairs own some of these desired abilities.
In addition to high accelerations and compliance, PAMs exhibit similarities to skeletal muscles in static and dynamic behavior~\cite{chou_measurement_1996,klute_mckibben_1999, tondu_mckibben_2006,chou_static_1994}.
\begin{figure}
	\centering
	\subfloat[]{\includegraphics[width=.49\columnwidth]{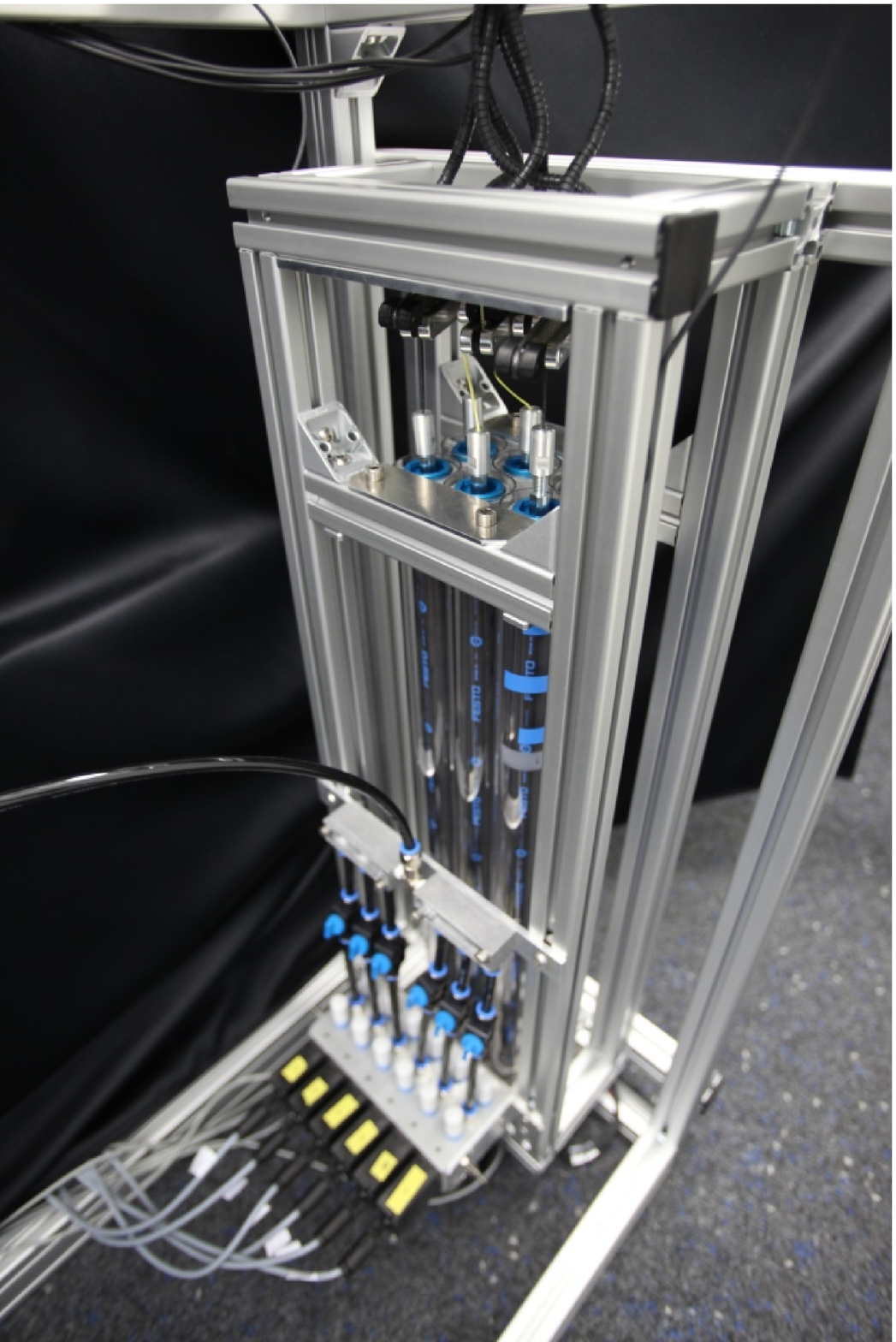}} 
	\hfill
	\subfloat[]{\includegraphics[width=.49\columnwidth]{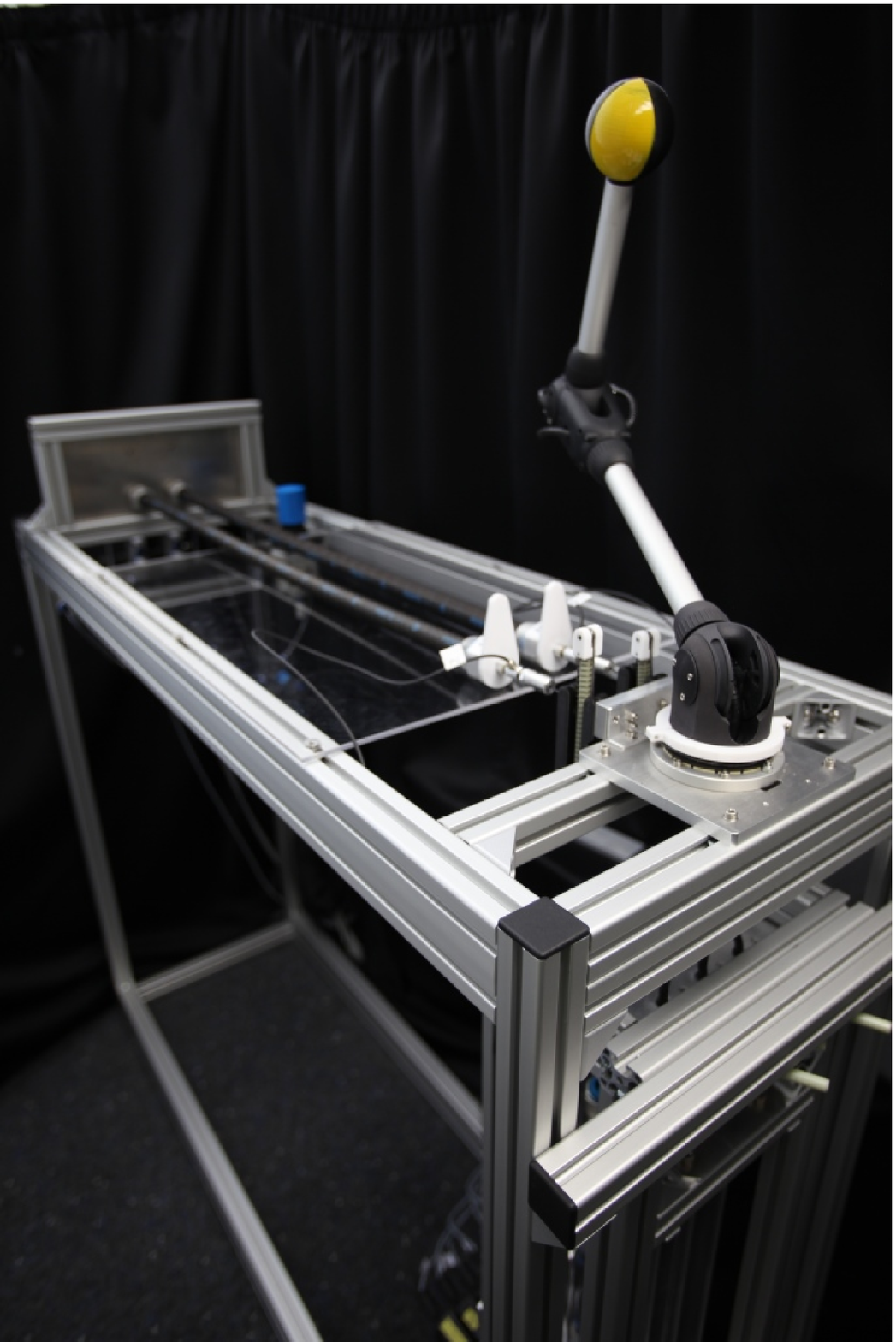}}
	\caption{Hardware components of our robot designed to keep friction low.
		(a) 6 PAMs are located directly below the Igus arm in order to pull the cables in the same direction as they exit the arm so that deflection is minimized.
		The necessary bending of the cables is realized by Bowden cables.
		(b) 2 PAMs actuating the first DoF are located on top of the base frame.
		They are longer~(\SI{1}{\meter}) than the other six PAMs~(\SI{0.6}{\meter}) due to the bigger radius of the first rotational DoF.}
	\label{fig:hardware}
\end{figure}
%
However, PAMs do not fully resemble the skeletal muscle. 
PAMs pull only along their linear axes and break when curled.
Muscle structures bending over bones like the deltoid muscles that connect the acromion with the humerus bone at the shoulder are hardly realizable.
Furthermore, biological muscles can be classified as wet-ware whereas PAMs suffer from additional friction when touching each other or the skeleton during usage. 
Thus, bi-articular configurations~(one PAM influences two DoFs) like the ones present in the human arm with seven DoFs are hard to realize.
Although it seems that PAMs are well suited to be attached directly to the joints instead of using cables due to their high power-to-weight ratio, this results in bigger moving masses and, thus, more non-linearities. 
\begin{figure*}[t]
	\centering
	\includegraphics[width = 6in]{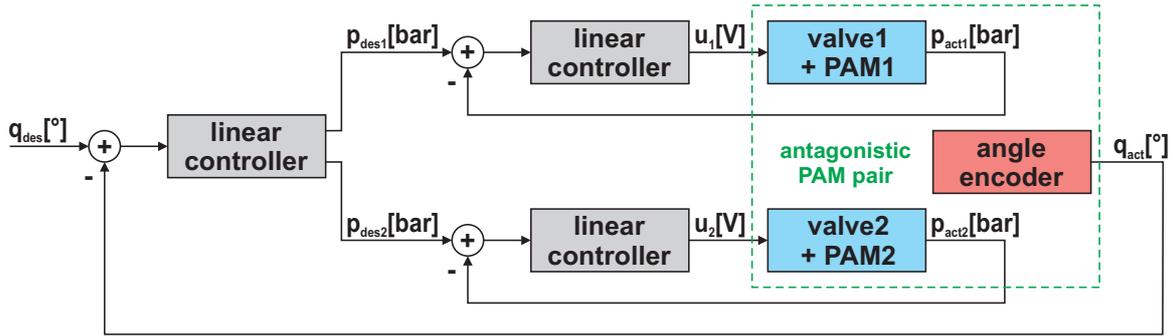}
	\caption{
		Schematic description of the position control loop for one PAM muscle pair.
		The absolute value of the output signal of the position control PID is assigned following the symmetrical co-contraction approach discussed in \sec{ssec:cocon}.
		The pressure within each PAM is governed by separate PIDs that set the input voltage to the proportional air flow valves.
		The sensor values are provided by Festo\texttrademark~pressure sensors and angle encoders.
	}
	\label{fig:cascade}
\end{figure*}

Many systems have been designed with the aim of reproducing the human anatomy using PAMs~(see \tab{tab:pamarm}).
Although such recent publications show good tracking performance of one PAM in position~\cite{zhong_one_2014,tondu_robust_2014,andrikopoulos_piecewise_2014}, using PAM-based systems with more DoFs for fast trajectory tracking appears to be less satisfactory.
The performance of PAM-actuated robots has thus been limited to slow movements compared to servo motor driven robots.
In \tab{tab:pamarm} we list, along with the existing PAM-actuated arms, the most complex (form and velocity) tracked trajectory in case it was mentioned.
For our purpose it is crucial that anthropomorphism does not degrade the ease to control the resulting arm. 

In this paper, we present a robot~(see \fig{fig:arm} and \fig{fig:hardware}) that fulfills our requirements while avoiding the problems of previous construction to achieve precise and \textit{fast} movements.
We illustrate the effectiveness of our hardware considerations by showing good results at tracking of slow trajectories with PIDs only. 
Additionally, we demonstrate high acceleration and velocity motions by applying step control signals to the PAMs. 
These motions surpass the peak velocity and acceleration of the Barrett WAM arm, that is used for robot table tennis~\cite{mulling_biomimetic_2011,huang_jointly_2016}, by a factor of 4x and 10x respectively while being able to sustain the mechanical stress. 
Another contribution of this paper is the tuning of control parameters using Bayesian optimization~(BO) \textit{without} any safety considerations.
Although previous papers employed BO on real robots~\cite{marco_automatic_2015,bansal_goal-driven_2017, antonova_sample_2016}, the applications have been limited to rather slow and safe motions whereas we even allow for unstable controllers during training as long as the motion in bounded~(see \sec{ssec:pid}). 
Our path is parallel to the sensible approach of taking safety directly into account, such as by means of constraints~\cite{berkenkamp_bayesian_2016}, where we enable safety through antagonistic actuation for high-acceleration tasks. 
Using the parameters learned with BO, we track - to the best of our knowledge - the fastest trajectory that has been tracked with a four DoF PAM-driven arm.
At last, we empirically show that choosing the appropriate co-contraction level is essential to achieve good control performance. 

We encourage other researchers to use our platform as a testbed for learning control approaches. 
We used off-the-shelf and affordable parts like PAMs by Festo\texttrademark, the robot arm by Igus and build the base using Item profiles.
All necessary documents to rebuild our system and videos of its performance can be found at~\url{http://musclerob.robotlearning.ai}. 
	\section{System Design to Generate and Sustain Highly Accelerated Movements}
	\label{sec:hardware}
	Using systems actuated by PAMs is a chance to improve performance at high-accelerations tasks on real robots by applying Machine Learning to tune low level controller.
On the other hand, such systems add additional control challenges. 
We identify the following key opportunities to ease control on such systems: 
1) avoiding friction between muscles,
2) avoiding contact between muscles and skeleton,
3) installing PAMs in the torso to decrease moving mass, 
4) minimal deflection of cables,
5) light-weight segments,
6) mostly independent DoFs.
These points guided the construction of our four DoF PAM-driven arm. 
\subsection{Igus\texttrademark~Robolink Lightweight Kinematics} 
To achieve high accelerations, it is generally desireable to have low moving masses.
At the same time, minimizing the weight also minimizes the non-linearities within a system~(especially the weight at the end-effector). 
Hence, we incorporate a light-weight tendon-driven arm by Igus\texttrademark~\cite{igus_robolink_2015} that has four DoFs and is actuated by eight PAMs~(two PAMs per DoF, \fig{fig:cascade}).
The arm has two rotational DoFs in each of the two joints and weighs less than \SI{700}{\gram} in total.
The first joint, which is fixed to the base, contributes little to the moving mass.
As a result, the PAM dynamics are dominant over the arm dynamics. 
In addition, it is driven by Dyneema tendons~($2\textrm{mm}$ diameter, tensile strength of \SI{4000}{\newton}) that allow fixing the PAMs in the base. 
Necessary deflections within the Igus\texttrademark arm are realized through Bowden cables.
They guide the cables within the arm almost without influencing each other and keep the length unchanged during movement.
As a result, crosstalking between DoFs due to cables is minimized.
Still, little cross-talking persists as the PAMs share the same air pressure supply as well as due to the non-zero moving mass.

\begin{figure}[t]
	\centering
	\subfloat[]{
		\centering\scriptsize%
		\input{stepFig}
		\label{fig:step}
	}
	\subfloat[]{
		\centering
		\includegraphics[width = 1.65in]{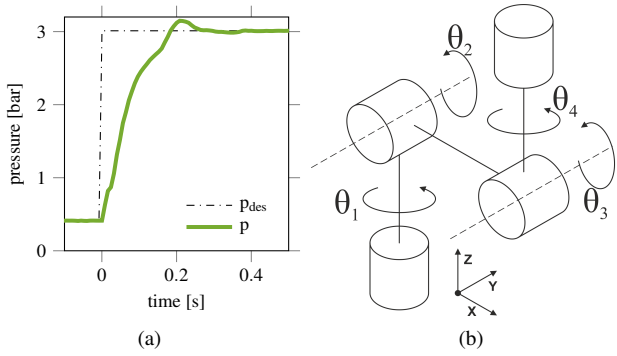}
	}
	\caption{
		(a) Pressure step response from minimum to maximum value of $3$~bar.
		The desired pressure value can be reached within approximately \SI{250}{\milli\second}.
		(b) Kinematic structure of the Igus\texttrademark~Robolink arm. 
		Two rotational DoFs are located at the base ($\theta_1$ and $\theta_2$) and two additional in the second joint ($\theta_3$ and $\theta_4$).
	}
	\label{fig:system}
\end{figure}

Cable-driven systems usually suffer from additional friction.
For this reason, the tendons are only minimally bended by our construction. 
All PAMs pull their respective tendons in the same direction as they exit the Igus\texttrademark~arm.
Two PAMs actuate the first rotational DoF in the base joint in horizontal orientation whereas the other 6 PAMs pull in vertical direction as can be seen in \fig{fig:hardware}a and b, respectively.  
The joint angles are measured by angular encoders with a resolution of approximately \SI{0.07}{\degree}.
The kinematic structure is depicted in \fig{fig:system}b.
%
\setlength{\figwidth }{0.35\columnwidth} 
\setlength{\figheight }{1\figwidth}   
\subsection{Reasons to Use Pneumatic Artificial Muscles}\label{ssec:pam}
Apart from a lightweight kinematics, generating high accelerations requires high forces. 
Hence, we use PAMs by Festo\texttrademark~ to actuate our robotic arm.
PAMs consist of an inner rubber tube surrounded by a braided weave composed of repeated and identical rhombuses. 
An increase in air pressure leads to a gain in diameter of the inner balloon. 
The double-helix-braided sheave transforms the axial elongation into a longitudinal contraction.
This contraction process can be fully characterized according to the radius of the inner tube and the braid angle.
The inner pressure plays the same role as the neuronal activation level of a biological muscle. 
The dynamics of both, PAMs and biological muscles, share some characteristics that are captured to some extent by the Hill muscle model~\cite{chou_measurement_1996,klute_mckibben_1999, tondu_mckibben_2006}
\begin{align}
(F+a)(V+b)=b(F_0+a)\,,
\end{align}
where $F$ and $V$ are the tension and contraction velocity of the muscle, $a$ and $b$ muscle-dependent empirical constants and $F_0$ the maximum isometric force generated in the muscle. 

In our robot, two \SI{1}{\meter} and six \SI{0.6}{\meter} PAMs, each with a diameter of \SI{20}{\milli\meter}, actuate four DoFs~($M=4$, two PAMs per DoF).
Each PAM can generate maximum forces of up to \SI{1200}{\newton} at \SI{6}{\bar}.
We limit the pressure to a maximum of \SI{3}{\bar} because the generated accelerations are sufficient, and to prolong the lifetime of the system. 
\fig{fig:step} shows that the maximum desired pressure can be reached within appr. \SI{250}{\milli\second}.
\app{ssec:software} describes the software framework in detail. 

The powerful actuation of PAMs comes with beneficial properties.
The high forces of PAMs better overcomes the resisting force of static friction.
Also, fast and catapult-like movements can be generated by pressurizing both PAMs and discharging one of them.
A similar kind of energy storage and release can also be found in human and primate arms in ballistic movements. 
At such high velocities, antagonistic muscle actuation can additionally be used to physically sustain high stress as well as ensure to stay within predefined joint ranges.
We pretension each PAM with an individual minimum pressures $\pmin{}\in\R{2M}$ so that a motion, that is opposing the direction of the torque of each muscle, is decelerated. 
Also, we set a maximum pressure to each PAM $\pmax{}\in\R{2M}$ so the torque generated due to $\pmin{}$ is sufficient to stop the motion. 
In this manner, our system does not exceeds fixed joint limits as the antagonist muscle to the motion is stretched and the torque produced by the agonist is sufficiently low. 
Consequently, any stable pressure trajectory - including step signals - are allowed within the predefined pressure ranges. 
In \sec{ssec:maxvelacc} we show that our system generates high accelerations within this pressure range. 
In contrast, robotic systems with servomotors must take care not to reach too high accelerations as the movement might not be decelerated fast enough before running into joint limits.

Despite advantageous properties, PAMs are hard to control which is the reason why they are not widely adopted.
PAMs pose hard challenges such as 1) the non-linear relationship between length, contraction velocity and pressure, 2) time-varying behavior~(as a result of dependencies on temperature and wearing) as well as 3) hysteresis effects~\cite{caldwell_control_1995,tondu_modelling_2012}.
These issues render modeling of PAM-driven systems challenging~\cite{buchler_control_2018}.
Thus, PAMs have been mainly applied due to their safety properties and high power-to-weight ratio for slow movements with the ability to carry heavy objects.
Using such a system as a testbed for learning control approaches is a promising direction.
	
	\section{Using Bayesian Optimization to Tune Control of Muscular System}
	\label{sec:background}
	Our system is capable of generating highly accelerated motions and allows to explore in such fast regimes.
The obvious next step is to automatically tune control parameters directly on the hardware.
We use Bayesian optimization~(BO) to learn to track a fast and hitting-like trajectory. 
This section introduces briefly explains the optimization method and PID control with feedforward compensation that is used to control this overpowered system.
Also, we adapt the symmetrical co-contraction approach so that co-contraction can be part of the tuning.
\subsection{Overpowered System Control}
\label{ssec:pid}
Tuning feedback controller for PAM-driven system is hard due to actuator delay, unobserved dependencies, non-linearities and hysteresis effects~\cite{buchler_control_2018, tondu_modelling_2012}.
Still, well-tuned linear Proportional Integral Derivative~(PID) controllers are often sufficient to track slow trajectories~(see \sec{ssec:pidtracking}).
The underlying control law
\begin{align}
\mathbf{u}_t=\mathbf{u}^\text{fb}_t+\mathbf{u}^\text{ff}_t\,,
\end{align}
consists of a feedback $\mathbf{u}^\text{fb}_t\in \R{M}$ and can be extended by a feedforward part $\mathbf{u}^\text{ff}_t\in \R{M}$ where $M$ represents the number of DoFs.
The PID feedback controller 
\begin{align}
	\mathbf{u}^\text{fb}_t=K^\text{P}\tilde{\mathbf{q}}_t+K^\text{D}\dot{\tilde{\mathbf{q}}}_t+K^\text{I}\tilde{\mathbf{q}}_t^s\,,
	\label{eq:compfb}
\end{align}
takes the position $\tilde{\mathbf{q}}_t=\mathbf{q}_t-\mathbf{q}^\text{des}_t\in\R{M}$, velocity $\dot{\tilde{\mathbf{q}}}_t=\dot{\mathbf{q}}_t-\dot{\mathbf{q}}^\text{des}_t\in\R{M}$ and integral errors $\tilde{\mathbf{q}}_t^s\in\R{M}$ as input where $[\tilde{\mathbf{q}}_t^s]_i=\int^t_0\tilde{q}_i(x)dx$.
The integral part compensates for steady-state errors. 
PIDs can be tuned by optimizing the elements of the feedback gain matrices $K^{x_\text{fb}}=\text{diag}(k^{x_\text{fb}}_1,\ldots,k^{x_\text{fb}}_M)$ with $x_\text{fb}\in \{\text{P,I,D}\}$.
The position feedback is always delayed by at least one cycle and hence subject to instabilities. 

Feedforward compensation 
\begin{align}
\mathbf{u}^\text{ff}_t=K^\text{pos}{\mathbf{q}}_t^\text{des}+K^\text{vel}\dot{\mathbf{q}}^\text{des}_t+K^\text{acc}\ddot{\mathbf{q}}_t^\text{des}\,,
\label{eq:compff}
\end{align}
%
instantly generates a control signal in response to the current desired joint position $\mathbf{q}^\text{des}_t$, velocity $\dot{\mathbf{q}}^\text{des}_t$ and acceleration $\ddot{\mathbf{q}}^\text{des}_t$.
In accordance to the feedback gain matrices, the feedforward gain matrices are also diagonal $K^{x_\text{ff}}=\text{diag}(k^{x_\text{ff}}_1,\ldots,k^{x_\text{ff}}_M)$ with ${x_\text{ff}}\in \{\text{pos,vel,acc}\}$.
Feedforward compensation has many beneficial properties.
First, feedforward terms can help to reduce the malicious effects of hysteresis. 
Second, feedforward terms do not affect the stability of the feedback part~\cite{brosilow_techniques_2002}, hence can be used purely to improve tracking performance. 
Third, tracking capabilities of pure feedback controllers are unavoidably degenerated for trajectories with high values of speed and accelerations.
Feedforward terms help in such situations as the reaction to a desired fast motion happens instantly and is not delayed by at least one cycle as in feedback control. 

The fact that the muscle dynamics are dominant over the rigid body dynamics has consequences for how we control the robot.
PAMs are dominant because they generate high forces while the arm is lightweight. 
For this reason, we decided to perform independent joint control, thus, we chose the gain matrices $K^{x_\text{c}}$ where ${x_\text{c}}\in \{\text{P,I,D,pos,vel,acc}\}$ to be diagonal although they generally have non-zero off-diagonal elements. 
Another consequence of the dominant PAM dynamics is that the non-linearities due to the rigid body dynamics are less effective.
Still, the PAM dynamics are non-linear and the linear dynamics assumption of PID controller with linear feedforward terms cannot be fulfilled. 
Fortunately, PIDs work well even for approximately linear systems. 
For this reason, we assume the parameters to be valid in the vicinity of a specific trajectory rather than for arbitrary desired motions.
In \sec{ssec:botrack}, we validate this claim by finding parameters that lead to good tracking performance on a fast trajectory.

Some sets of parameters for PID controllers lead to instabilities. 
Instabilities can cause damages by 1) running the links into their particular joint limits with high velocities and make the robot hit itself or by 2) creating high internal forces that break parts inside the robot such as connections to cables.
For the latter case, we show in \sec{ssec:maxvelacc} that our system sustains step pressure signals that generate a highly accelerated and fast motion. 
The high internal forces created by such a motion did not damage any internal parts, most probably because the PAMs are backdrivable. 
The first case is more complicated: Our system does not break for periodic motions caused by instable controllers as long as the motion is bounded, a term we coin as \textit{bounded instabilities}.
In other words, the control signal does not excite the resonance frequency of the motion and add energy with every period.  
The bound for the periodic motion can be as wide as the allowed joint limits for each DoF. 
From our experience, it is sufficient to adjust an upper limit to the elements of the feedback gain matrices.
If an unbounded instability occurs, the robot can be stopped by releasing the air from the PAMs and manually holding the robot. 
The low inertia due to the low moving masses as well as the backdrivable PAMs enables to proceed in this manner. 
This procedure is not possible on traditional motor-driven systems as the higher inertia can cause higher forces at impact and the motors are usually not backdrivable. 
Additionally, traditional systems would break due to high internal forces generated at such behaviors. 
In \sec{ssec:botrack} we tune PID parameter with BO without further safety considerations on a fast trajectory while allowing bounded instabilities.
%
\setlength{\figwidth }{0.9\columnwidth} 
\setlength{\figheight }{.309\figwidth}   
\begin{figure}[t]
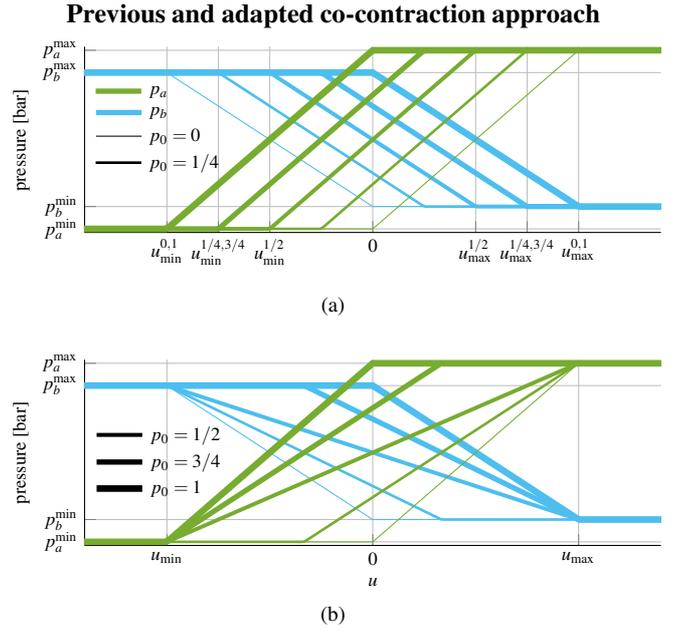

	\centering
	\textbf{Previous and adapted co-contraction approach}\par\medskip
	\vspace{-.5cm}
	\subfloat[]{
		\centering\scriptsize%
		\input{deltap}
		\label{sfig:deltap}
	}\\
	\subfloat[]{
		\centering\scriptsize%
		\input{deltapcorr}
		\label{sfig:deltapcorr}
	}
	\caption{
		Approach to assign both pressures $p_a$ and $p_b$ of an antagonistic PAM pair from a scalar control signal $u$, hence, converting from a MISO into a SISO system. 
		The thickness of the lines indicate different $p_0$ values defined in \eq{eq:dp} and \eq{eq:dpcorr}. 
		The two colors represent  $p_a$ and $p_b$ respectively. 
		The sum of $p_a$ and $p_b$ increases with increasing $p_0$, thus, increasing the stiffness in the antagonistic PAM pair. 
		(a) Symmetrical pressure approach from \eq{eq:dp} with additional saturation to keep $p_a$ and $p_b$ within the allowed ranges. 
		The control range $[u_\text{min},u_\text{max}]$ that effectively changes at least $p_a$ or $p_b$ changes for varying $p_0$ where the superscript indicates $p_0$ for the respective control range~(e.g. $u_\text{min}^{0,1}$ stands for lower range limit for $p_0=0$ and $p_0=1$). 
		(b) Approach that corrects for changing effective control ranges for varying $p_0$ by adapting the slope within $[u_\text{min},u_\text{max}]$ with $c$~(see \eq{eq:dpcorr}).
	}
	\label{fig:deltap}
\end{figure}
\noindent
\subsection{Adapted Symmetrical Co-contraction Approach}
\label{ssec:cocon}
An antagonistic PAM pair is a multi-input-single-output-system~(MISO) where both PAMs $p_a$ and $p_b$ influence the joint angle $q$. 
In contrast, a controller for a traditionally motor-driven robot outputs a scalar control signal $u$ for each DoF.
In our case, this scalar control signal has to map to both desired pressure for both PAMs $p^{\text{des}}_a$ and $p^{\text{des}}_b$ of an antagonistic pair to form a single input single output~(SISO) system.
One way is to assign the control signal in opposing directions to each PAM
\begin{align}
	p_{x}&=p_0 \pm u\,,\label{eq:dp}
\end{align}
as suggested in \cite{tondu_seven-degrees--freedom_2005} where $x\in\{a,b\}$.
Here, the assumption is that the joint angle $q$ of each DoF increases with rising $p_a$ and falling $p_b$ and vice versa. The co-contraction parameter $p_0$ correlates with the stiffness of the corresponding DoF.
However, the input range for the control signal $[u_\text{min},u_\text{max}]$ for which at least one of the pressures $p_a$ and $p_b$ change depends on the value of $p_0$ as can be seen in \fig{sfig:deltap}.
A fixed input range is essential in case $p_0$ should be optimized for control next to the elements of the gain matrices from \eq{eq:compfb} and \eq{eq:compff}.
We extend this approach by first allowing the scalar control signal $u$ to be only within $[-1,1]$ and, secondly, linearly map the resulting number to an allowed pressure range
\begin{align}
	p_{x}=(p_{x}^\text{max}-p_{x}^\text{min})\big(p_0 \pm c~\textrm{sat}(u,-1,1)\big)+p_x^\text{min}\label{eq:dpcorr}\,,
\end{align}
where $p_{x}^\text{max}$ and $p_{x}^\text{min}$ with $x\in\{a,b\}$ are the individual maximal and minimal pressures.
The saturation function $\text{sat}(\cdot,l_\text{min},l_\text{max})$ is depicted in \fig{fig:sat} with $l_\text{min}$ and $l_\text{max}$ being the lower and upper threshold. 
\setlength{\figwidth }{0.4\columnwidth} 
\setlength{\figheight }{.618\figwidth}   
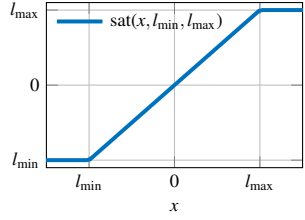
\begin{wrapfigure}{r}{0.48\columnwidth}
	\centering\scriptsize%
	\hspace{-1cm}
%
%
\definecolor{mycolor1}{rgb}{0.00000,0.44700,0.74100}%
\begin{tikzpicture}

\begin{axis}[%
width=0.951\figwidth,
height=\figheight,
at={(0\figwidth,0\figheight)},
scale only axis,
xmin=-1.5,
xmax=1.5,
xtick={-1,0,1},
xticklabels={{$l_\textrm{min}$},{0},{$l_\textrm{max}$}},
xlabel style={font=\color{white!15!black}},
xlabel={$x$},
ymin=-1.1,
ymax=1.1,
ytick={-1,0,1},
yticklabels={{$l_\textrm{min}$},{0},{$l_\textrm{max}$}},
axis background/.style={fill=white},
xmajorgrids,
ymajorgrids,
legend style={at={(0.03,0.97)}, anchor=north west, legend cell align=left, align=left, draw=white!15!black},
ylabel near ticks,
xlabel near ticks,
legend style={at={(0,1)},anchor=north west,legend cell align=left,align=left,fill=none,draw=none}
]
\addplot [color=mycolor1, line width=1.5pt]
  table[row sep=crcr]{%
-2	-1\\
-1.95959595959596	-1\\
-1.91919191919192	-1\\
-1.87878787878788	-1\\
-1.83838383838384	-1\\
-1.7979797979798	-1\\
-1.75757575757576	-1\\
-1.71717171717172	-1\\
-1.67676767676768	-1\\
-1.63636363636364	-1\\
-1.5959595959596	-1\\
-1.55555555555556	-1\\
-1.51515151515152	-1\\
-1.47474747474747	-1\\
-1.43434343434343	-1\\
-1.39393939393939	-1\\
-1.35353535353535	-1\\
-1.31313131313131	-1\\
-1.27272727272727	-1\\
-1.23232323232323	-1\\
-1.19191919191919	-1\\
-1.15151515151515	-1\\
-1.11111111111111	-1\\
-1.07070707070707	-1\\
-1.03030303030303	-1\\
-0.98989898989899	-0.98989898989899\\
-0.949494949494949	-0.949494949494949\\
-0.909090909090909	-0.909090909090909\\
-0.868686868686869	-0.868686868686869\\
-0.828282828282828	-0.828282828282828\\
-0.787878787878788	-0.787878787878788\\
-0.747474747474747	-0.747474747474747\\
-0.707070707070707	-0.707070707070707\\
-0.666666666666667	-0.666666666666667\\
-0.626262626262626	-0.626262626262626\\
-0.585858585858586	-0.585858585858586\\
-0.545454545454545	-0.545454545454545\\
-0.505050505050505	-0.505050505050505\\
-0.464646464646465	-0.464646464646465\\
-0.424242424242424	-0.424242424242424\\
-0.383838383838384	-0.383838383838384\\
-0.343434343434343	-0.343434343434343\\
-0.303030303030303	-0.303030303030303\\
-0.262626262626263	-0.262626262626263\\
-0.222222222222222	-0.222222222222222\\
-0.181818181818182	-0.181818181818182\\
-0.141414141414141	-0.141414141414141\\
-0.101010101010101	-0.101010101010101\\
-0.0606060606060606	-0.0606060606060606\\
-0.0202020202020201	-0.0202020202020201\\
0.0202020202020203	0.0202020202020203\\
0.0606060606060606	0.0606060606060606\\
0.101010101010101	0.101010101010101\\
0.141414141414141	0.141414141414141\\
0.181818181818182	0.181818181818182\\
0.222222222222222	0.222222222222222\\
0.262626262626263	0.262626262626263\\
0.303030303030303	0.303030303030303\\
0.343434343434343	0.343434343434343\\
0.383838383838384	0.383838383838384\\
0.424242424242424	0.424242424242424\\
0.464646464646465	0.464646464646465\\
0.505050505050505	0.505050505050505\\
0.545454545454545	0.545454545454545\\
0.585858585858586	0.585858585858586\\
0.626262626262626	0.626262626262626\\
0.666666666666667	0.666666666666667\\
0.707070707070707	0.707070707070707\\
0.747474747474747	0.747474747474747\\
0.787878787878788	0.787878787878788\\
0.828282828282828	0.828282828282828\\
0.868686868686869	0.868686868686869\\
0.909090909090909	0.909090909090909\\
0.949494949494949	0.949494949494949\\
0.98989898989899	0.98989898989899\\
1.03030303030303	1\\
1.07070707070707	1\\
1.11111111111111	1\\
1.15151515151515	1\\
1.19191919191919	1\\
1.23232323232323	1\\
1.27272727272727	1\\
1.31313131313131	1\\
1.35353535353535	1\\
1.39393939393939	1\\
1.43434343434343	1\\
1.47474747474747	1\\
1.51515151515152	1\\
1.55555555555556	1\\
1.5959595959596	1\\
1.63636363636364	1\\
1.67676767676768	1\\
1.71717171717172	1\\
1.75757575757576	1\\
1.7979797979798	1\\
1.83838383838384	1\\
1.87878787878788	1\\
1.91919191919192	1\\
1.95959595959596	1\\
2	1\\
};
\addlegendentry{sat($x,l_\textrm{min},l_\textrm{max}$)}

\end{axis}
\end{tikzpicture}%
	\caption{
		Saturation function
	}
	\label{fig:sat}
	\vspace{-12pt}
\end{wrapfigure}
\noindent
The saturation function that keeps $u$ withing the range $[-1,1]$ in \eq{eq:dpcorr} ensures that the computed desired pressures stay within the allowed ranges. 
Additionally, we added a correction parameter $c=0.5-\text{sat}(|p_0-0.5|,0,0.5)$ with the absolute value $|\cdot|$, to create different slopes depending on the value of $p_0$.
\fig{sfig:deltapcorr} depicts our corrected solution.
\subsection{Bayesian Optimization}
\label{ssec:bo}
Bayesian optimization~(BO) is a zero-order optimization technique~\cite{mockus_bayesian_1989,jones_taxonomy_2001,shahriari_taking_2016} that aims at finding the global minimum 
\begin{align}
	\mathbf{x}^*=\argmin_{\mathbf{x}}f(\mathbf{x})\,,
\end{align}
of an unknown function~$f:\R{D} \rightarrow \mathbb{R}$ with inputs $\mathbf{x}\in\R{D}$.
BO operates in a black-box manner as the function is not required in analytical form but rather is modeled as a response surface $\tilde{f}$ from samples collected in a dataset  $\mathcal{D}=\{(\mathbf{x}_i,f(\mathbf{x}_i))|i=0,1,\dots N-1\}$ of the input parameters $\mathbf{x}_i \in \mathbb{R}^D$ and the resulting function evaluation $f(\mathbf{x}_i)\in \mathbb{R}$.
Often probabilistic regression techniques are incorporated to handle noisy observations in a principled way, take model uncertainties into account and allow to integrate domain knowledge using priors. 
Among other methods, Gaussian Processes~(GPs~\cite{rasmussen_gaussian_2006}) are widely used. 
A GP is a distribution over functions where the conditional posterior distribution is Gaussian 
\begin{align}
	p(\tilde{f}|X,\mathbf{y},\mathbf{x}_*)=\mathcal{N}(\mu,\sigma^2)\,,\label{eq:posterior}
\end{align} 
with mean and variance
\begin{align}
	\mu(\mathbf{x}_*)&=\mathbf{k}_*^T(K+\sigma_n^2I)^{-1}\mathbf{y}\,,\\
	\sigma^2(\mathbf{x}_*)&=k_{**}-\mathbf{k}_*^T(K+\sigma_n^2I)^{-1}\mathbf{k}_*\,,
\end{align}
where $X\in \mathbb{R}^{N\times D}$ is a design matrix with each row being the n-th training input $\mathbf{x}_n^T\in\R{1\times D}$, $\mathbf{y}\in\R{D}$ are the target values, $[K]_{i,j}=k(\mathbf{x}_i,\mathbf{x}_j)$, $[\mathbf{k}_*]_i=k(\mathbf{x}_i,\mathbf{x}_*)$, $k_{**}=k(\mathbf{x}_*,\mathbf{x}_*)$ and $I$ is the identity matrix.
The function $k(\mathbf{x}_a,\mathbf{x}_b)$ is a kernel that represents the correlation between two data points $\mathbf{x}_a$ and $\mathbf{x}_b$.  
Here, we consider the Mat\'{e}rn $5$ kernel with Automatic Relevance Determination~(ARD)
\begin{align}
	k(\mathbf{x}_a,\mathbf{x}_b)=\sigma_f\bigg(1+\sqrt{5}r+\frac{5}{3}r^2\bigg)\exp(-\sqrt{5}r)+\sigma_n\delta_{a,b}\,,
\end{align}
where $r^2=\sum_{d=0}^{D-1}(x_{ad}-x_{bd})^2/l^2_d$, $l^2_d$ is an individual lengthscale for each input dimension and $\sigma_n$ and $\sigma_f$ are the noise and the signal variances. 

In every iteration BO chooses the next query point $\mathbf{x}'$ according to a surrogate function, also called activation surface
\begin{align}
	\mathbf{x}^*=\argmin_{\mathbf{x}}\alpha(\mathbf{x})\,,
\end{align}
rather than optimizing the response surface $\tilde{f}$ directly.
Different acquisition functions exist~\cite{shahriari_taking_2016} that focus on various criteria such as \emph{improvement}~(probability of improvement and expected improvement), \emph{optimism}~(upper confidence bound) or \emph{information}~(Thompson sampling and entropy search) to name just a few. 
All of them aim at balancing exploration and exploitation to maximize sample efficiency by taking advantage of the full posterior distribution from \eq{eq:posterior}
\begin{align}
	\alpha(\mathbf{x})=\mathbb{E}_{p(\tilde{f}|\mathcal{D},\mathbf{x})}[U(\mathbf{x},\tilde{f}(\mathbf{x}))]\,,
\end{align}
where $U(\mathbf{x},\tilde{f}(\mathbf{x}))$ defines the various quality criteria mentioned above.
In particular, we incorporate \emph{expected improvement}~(EI)
\begin{align}
	U(\mathbf{x},\tilde{f}(\mathbf{x}))=\max(0,\tilde{f}'-\tilde{f}(\mathbf{x}))\,,
\end{align}
%
where $\tilde{f}'$ is the minimal value of $\tilde{f}$ observed so far. 
In the context of control parameter tuning, the inputs $\mathbf{x}$ correspond to the control parameter $\tta{}$ and function evaluations $\tilde{f}(\mathbf{x})$~(targets $\mathbf{y}$ to the GP)~are measurements of the control performance $\mathcal{L}$~(we define both in \sec{ssec:autotune}). 
For a comparison of BO approaches to robotics see \cite{calandra_bayesian_2016}.
\subsection{Automatic Tuning of PID Parameter for Antagonistic PAMs}
\label{ssec:autotune}
It is desirable to automatically tune control parameters directly on the real hardware. 
A significant concern is the possibility to cause damage to the robot or surrounding objects. 
Poorly chosen control parameter can cause too fast motions that cannot be decelerated in time.
The cause can be fast changing control signal, such as step signals, or instabilities.
\begin{algorithm}[t]
	\caption{Bayesian Optimization Parameter Tuning of a PID with feedforward compensation and co-contraction}
	\label{alg:bo}
	\begin{algorithmic}[1]
		\Procedure{BOParaTuning}{$N,\tman{},\tlim{},\mathbf{w},\boldsymbol{\tau}^\text{des}$}
		\For{$i_\text{dof}=1\dots M}$ ~\textcolor{gray}{\textit{//~all DoFs}}
		\State $\mathcal{D}\gets \emptyset$
		\State $\tta{i_\text{dof}}\gets \text{uniformrand()}$
		\For{$i_\text{other} \ne i_\text{dof}$} ~\textcolor{gray}{\textit{//~all not current DoFs}}
		\If{$i_\text{other}>i_\text{dof}$}
		\State $\tta{i_\text{other}}\gets \tman{i_\text{other}}$
		\Else
		\State $\tta{i_\text{other}}\gets \topt{i_\text{other}}$
		\EndIf
		\EndFor
		\For{$i_\text{it}=1\ldots N_\text{it}$}
		\State $\boldsymbol{\tau}~\gets$ track($\tta{i_\text{dof}},\boldsymbol{\tau}^\text{des}$) 
		\State $\mathcal{L}_\text{pos}\gets (\mathbf{q}-\mathbf{q}^\text{des})^T(\mathbf{q}-\mathbf{q}^\text{des})$
		\State $\mathcal{L}_\text{vel}\gets (\dot{\mathbf{q}}-\dot{\mathbf{q}}^\text{des})^T(\dot{\mathbf{q}}-\dot{\mathbf{q}}^\text{des})$
		\State $\mathcal{L}\gets w_\text{pos}\mathcal{L}_\text{pos}+w_\text{vel}\mathcal{L}_\text{vel}+w_\text{acc}\mathcal{L}_\text{act}(\mathbf{p})$
		\State $\mathcal{D} \gets [\mathcal{D},(\boldsymbol{\theta}_{i_\text{dof}},\mathcal{L})]$
		\State $\boldsymbol{\theta}_{i_\text{dof}}\gets \text{BO}(\mathcal{D},\tlim{})$
		\EndFor
		\State $\topt{i_\text{dof}}\gets \tta{i_\text{dof}}$
		\EndFor
		\EndProcedure
	\end{algorithmic}
\end{algorithm}
\setlength{\figwidth }{1.9\columnwidth} 
\setlength{\figheight }{.309\figwidth}  
\begin{figure*}[t]
	\centering
	\textbf{Tracking performance on slow trajectories}\par\medskip
	\scriptsize%
	\hspace{-1cm}
	\input{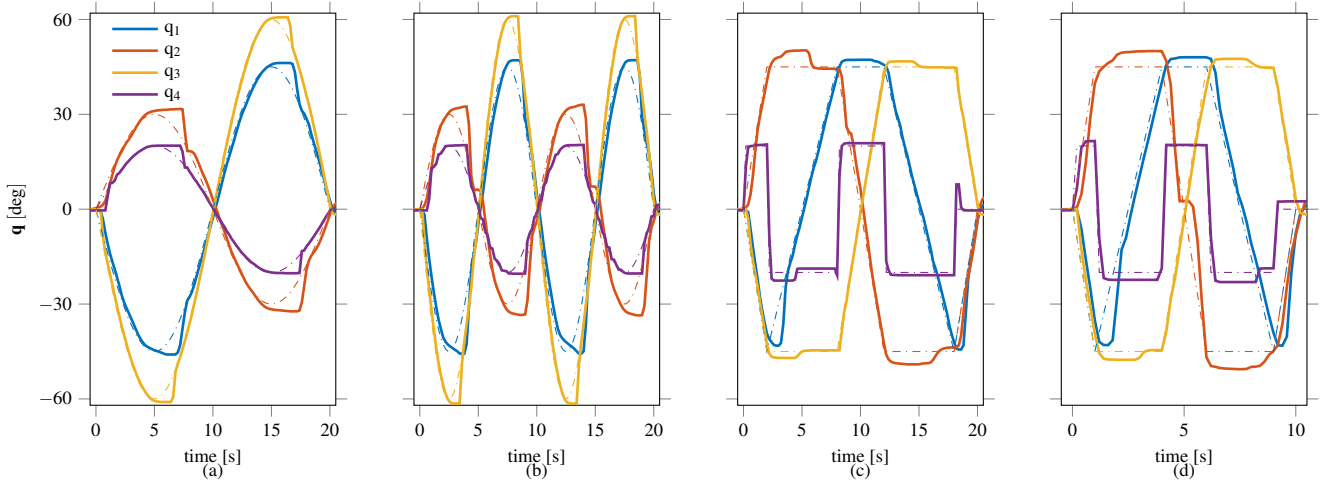}
	\caption{
		The tracking performance shows satisfactory results using a manually-tuned PID controller. 
		For rapid changes in reference signals, some overshoots are visible that cannot be compensated.
		For smooth changes as in (b) the trajectory is tracked sufficiently well indicating that our construction considerations from \sec{sec:hardware} eased the control of the robot. 
		(a) Sinusoidal reference with f=\SI{0.05}{\hertz}. 
		(b) Sinusoidal reference with f=\SI{0.1}{\hertz}. 
		(c) Truncated ramp reference for DoF one to three and rectangular reference for DoF four. 
		(d) Same reference as in (c) but twice as fast.
	}
	\label{fig:tracking}
\end{figure*}
On the other hand, high-acceleration robotics tasks require automatic tuning as fast motions are inherently harder to control while being even more susceptible to produce damages. 
Using our system, we can both generate highly accelerated movements and assure that such motions do not cause damage to the system as described in \sec{ssec:pid}.
Consequently, learning control algorithms can experience such explosive motions and incorporate this information rather than avoiding it. 

To illustrate this point, we aim at automatically tuning the PID control framework from \sec{ssec:pid} using the BO approach described in \sec{ssec:bo}.
We do so with no further safety considerations than to assume predefined pressure ranges $\plim{}\in \mathcal{R}^{2M\times 2}$ and $[\plim{}]_{m}=(p^\text{max}_{m},p^\text{min}_{m})$ that assure the robot to not hit its base.
Additionally, we set limits on the parameters $\boldsymbol{\theta}^\text{lim}$ that ensure that the maximal instabilities stay bounded~(see \sec{ssec:pid}).
The system still reaches high velocities and accelerations within these ranges as described in \sec{ssec:maxvelacc}.

A straightforward approach is to optimize the feedback $\mathbf{k}_i^\text{fb}=[k_i^\text{P},k_i^\text{D},k_i^\text{I}]$ and feedforward terms $\mathbf{k}_i^\text{ff}=[k_i^\text{pos},k_i^\text{vel},k_i^\text{acc}]$ for each DoF $i$.
Additionally, we tune the co-contraction parameter $p_0$ from \eq{eq:dpcorr}. 
It fundamentally changes how pressures are assigned based on the control signal and hence influences the system's characteristics. 
Hence, we optimize $\boldsymbol{\theta}_i=[\mathbf{k}_i^\text{fb},\mathbf{k}_i^\text{ff},p_{0,i}]$ for each DoF $i$ separately. 
We employ our adapted co-contraction approach from \eq{eq:dpcorr} as it enables $p_0$ to be part of the tuning~(the input range changes using \eq{eq:dp} but not using \eq{eq:dpcorr}).
While optimizing one of the DoFs, the others are either controlled by the previously optimized parameters  $\topt{i}$ or by manually tuned parameters $\tman{i}$ that are depicted in \fig{fig:botrack}.
The tracking using $\tman{}$ are similar to the tacking with $\topt{}$.
Hence, the influence of the other DoFs on the currently tuned DoF stays approximately constant throughout the optimization procedure. 

Control performance is hard to measure with a scalar quantity and is inherently multiobjective.
Taking inspiration from the LQR framework, we define the losses on position control error
\begin{align}
\Lp=(\mathbf{q}-\mathbf{q}^\text{des})^T(\mathbf{q}-\mathbf{q}^\text{des})\,,
\label{eq:lp}
\end{align}
and velocity control error
\begin{align}
\Lv=(\dot{\mathbf{q}}-\dot{\mathbf{q}}^\text{des})^T(\dot{\mathbf{q}}-\dot{\mathbf{q}}^\text{des})\,,
\label{eq:lv}
\end{align}
over a given desired trajectory $\boldsymbol{\tau}^\text{des}=[\mathbf{q}^\text{des},\dot{\mathbf{q}}^\text{des}]\in \mathbb{R}^{N_t\times 2}$ which is different from $\mathbf{q}_t^\text{des},\dot{\mathbf{q}}_t^\text{des}\in \mathbb{R}^\text{M}$ from \eq{eq:compff} that indicate joint angles and joint velocities of all DoFs at time $t$.
The goal is also to allow any curvature of the pressures $p_a$ and $p_b$ of PAMs $a$ and $b$ of an antagonistic pair~(also step signals) but keep them inside predefined pressure ranges $\plim{}$ over time. 
This property can be encoded by keeping the control signal $u$ within $[-1,1]$ using \eq{eq:dpcorr}.
Hence, we additionally generate an action loss
\begin{align}
\La=\begin{cases}
0 & \text{if } -1\leq u \leq 1\\
|u|-1& \text{otherwise} 
\end{cases}\,,
\label{eq:la}
\end{align}
if $u$ is out of its allowed range $[-1,1]$.
We scalarize to a single loss by computing a weighted sum
\begin{align}
\mathcal{L}=\sum_{k=\{\text{pos},\text{vel},\text{act}\}}w_k\mathcal{L}_k\,,
\end{align}
which enables us to reuse samples of the losses by saving them separately and reweighting with different $\mathbf{w}=[w_\text{pos},w_\text{vel},w_\text{act}]$.
A pseudocode of our BO approach is represented in \alg{alg:bo}.
\setlength{\figwidth }{1.9\columnwidth} 
\setlength{\figheight }{0.618\figwidth}   
\begin{figure*}[t]
	\centering
	\textbf{Ballistic movements}\par\medskip
	\scriptsize%
	\hspace{-1cm}
	\input{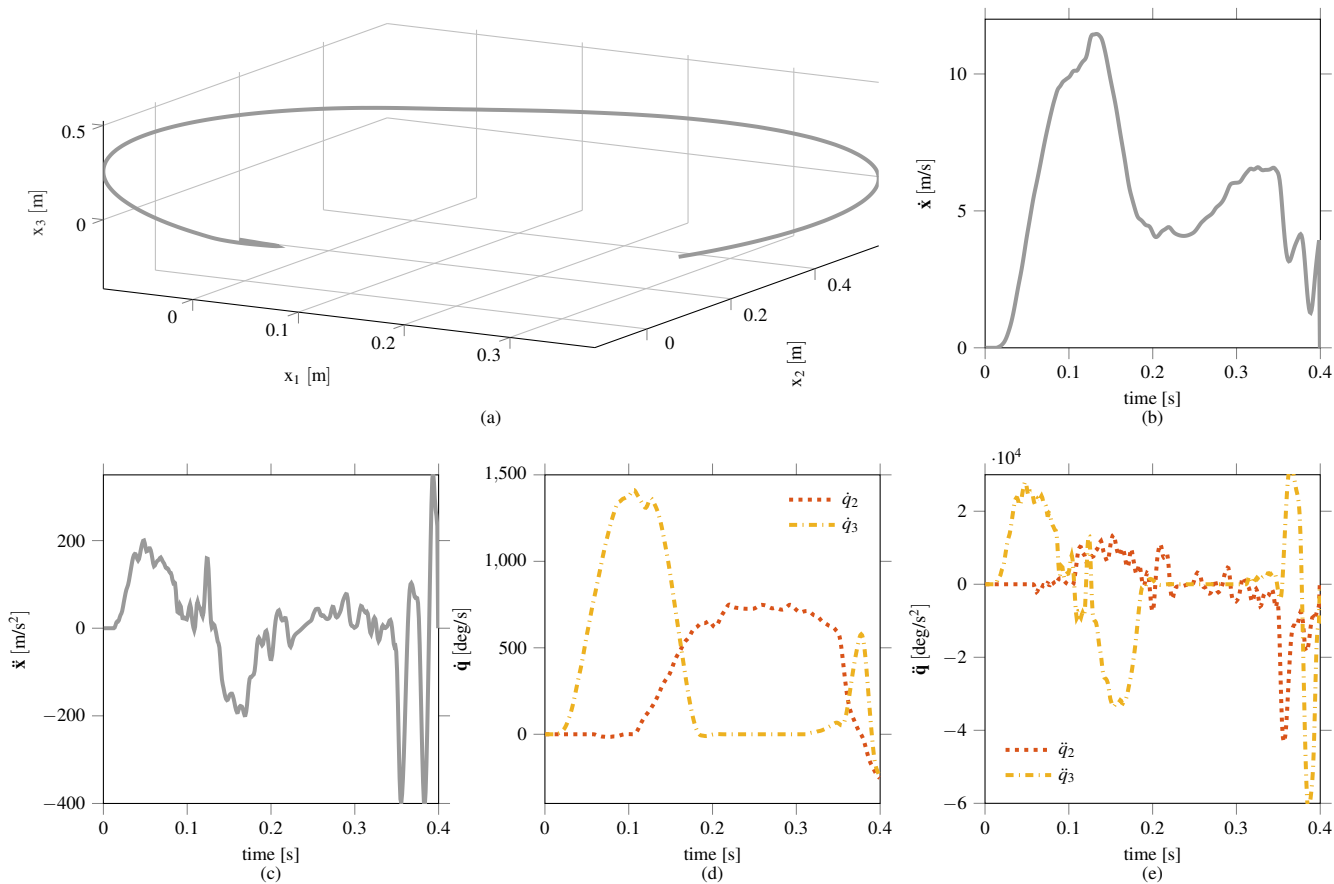}
	\caption{High velocity and acceleration motions in task and joint space.
		The pressures of the PAMs of DoF two and three were switched from minimal desired pressure to the maximum or vice versa. This switching generated a highly accelerated and fast motion.
		Our system sustained the stress generated from such motions which indicates that this system can be used to learn directly on the real hardware. 
		(a) Trajectory of the end-effector in task space.
		(b) Velocity profile along the trajectory in (a). 
		Maximum value is \SI{12}{\meter\per\second}.
		(c) Acceleration profile along the trajectory in (a).
		The maximum value reaches up to \SI{200}{\meter\per\second\squared}.
		(d) Angular velocity profile for rotational DoFs two and three.
		DoF three is faster as it has to accelerate less weight than DoF two.
		The maximum value of about \SI{1400}{\degree\per\second} is reached with DoF three.
		(e) Angular accelerations show a maximum of approximately \SI{28000}{\degree\per\second\squared}.
		\TODO{rotate $\mathbf{x}_2$ in (a)}}
	\TODO{find title style={below=46ex,font=\normalfont} and exchange for title style={below=47ex,font=\normalfont}}
	\label{fig:highvelacc}
\end{figure*}

In multi-objective optimization, one single set of parameters does not optimize all objectives at the same time, either due to the optimization being stopped early or the objectives contradict each other. 
The two losses $\Lp$ and $\Lv$ that we use here~(omitting the action loss $\La$) contradict each other as the linear controller from \eq{eq:compff} is not capable of achieving perfect tracking although theoretically $\Lp$ and $\Lv$ are zero once one of the objectives is zero.
Expressing optimality in the multi-objective case is done by calculating the Pareto front~(PF).
This set consists of points of non-dominated parameters where parameters $\tta{1}$ dominate parameters $\tta{2}$ if
\begin{align}
\tta{1}\succ \tta{2}
\begin{cases}
\forall~i=[1,N]:\mathcal{L}_i(\tta{1})\leq \mathcal{L}_i(\tta{2})\\
\exists~j=[1,N]:\mathcal{L}_j(\tta{1})< \mathcal{L}_j(\tta{2})\,,
\end{cases}
\label{eq:pareto}
\end{align}
which, in other words, means that $\tta{1}$ is strictly better in at least one objective compared to $\tta{2}$ and not worse in all other objectives.

\TODO{mention other control architectures}
\TODO{other loss functions?}

	
	
	\section{Experiments and Evaluations}
	\label{sec:experiments}
	Having derived our approach to automatically tune control parameters for our developed system, we now perform experiments to demonstrate its feasibility. 
First, we show that our construction considerations make it possible to track slow trajectories with PIDs only.  
In the second experiment, we demonstrate high velocity and acceleration motions to underline the ability of the arm to be used for hitting and generate catapult-like motions while avoiding damages at such paces. 
In the last experiment, both preceding experiments are combined by learning control~(PID with feedforward terms and co-contraction) directly on the real hardware without additional safety considerations on a fast and hitting-like trajectory.
\subsection{Control of Slow Movements}\label{ssec:pidtracking}
\setlength{\figwidth }{0.4\columnwidth} 
\setlength{\figheight }{.618\figwidth} 
\begin{figure*}
	\centering
	\textbf{Visualization of learned hitting motion}\par\medskip
	\vspace{-.5cm}
	\subfloat[time~$=\SI{1}{\second}$]{
		\centering\scriptsize%
		\includegraphics[scale=0.085]{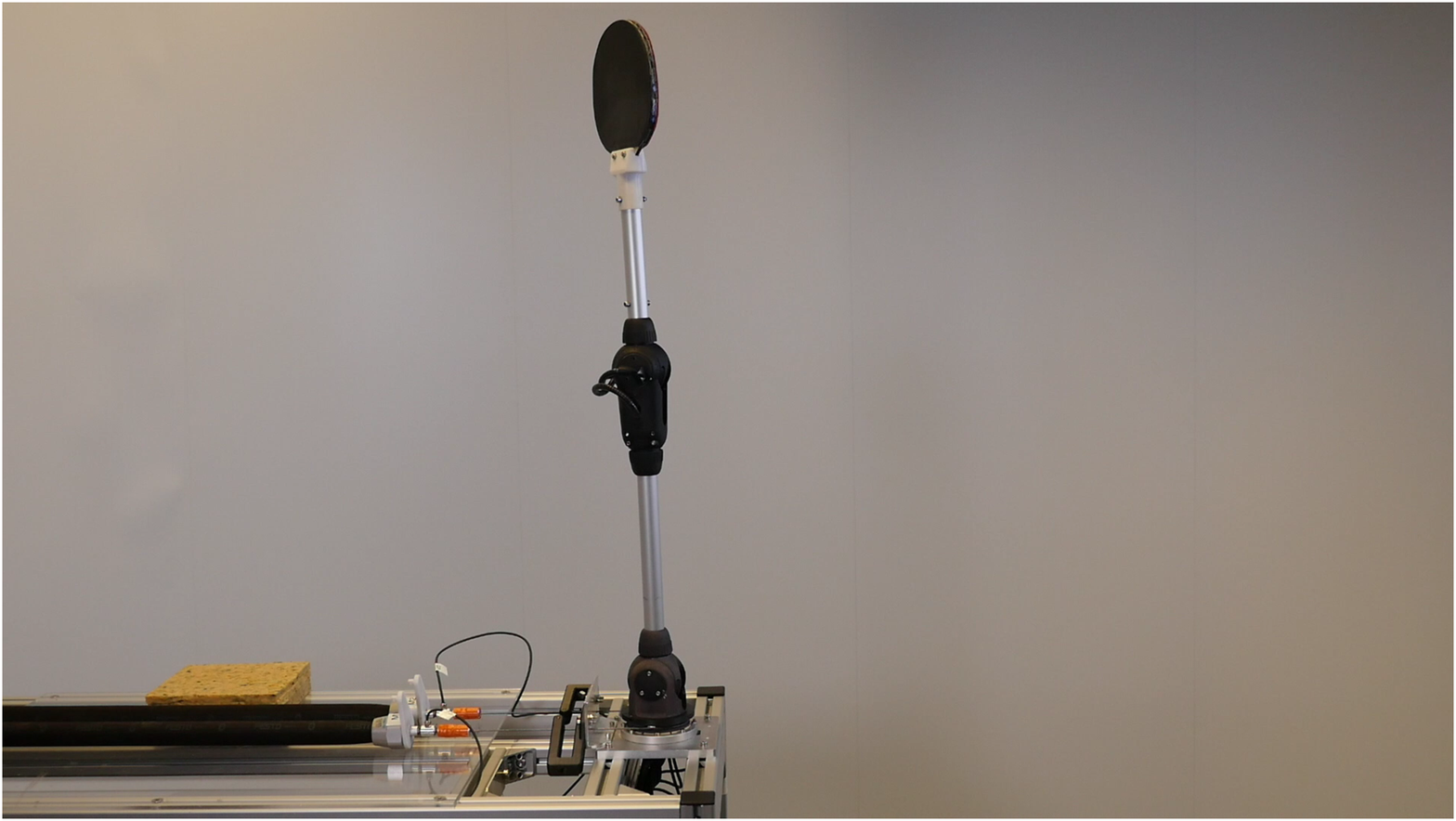}
	}
	\subfloat[time~$=\SI{1}{\second}\ldots\SI{3}{\second}$]{
		\centering\scriptsize%
		\includegraphics[scale=0.085]{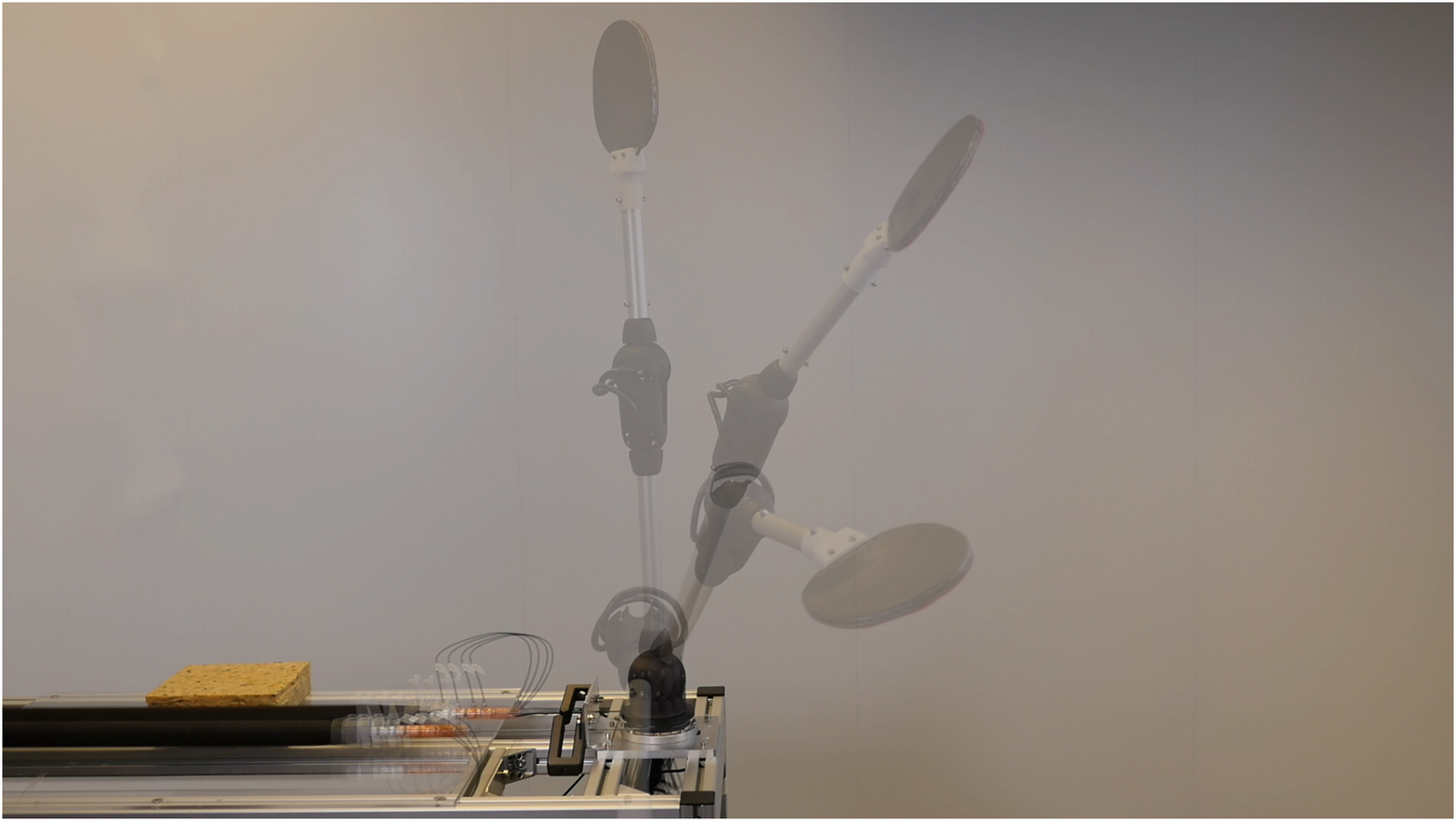}
	}
	\subfloat[time~$=\SI{3}{\second}\ldots\SI{4}{\second}$]{
		\centering\scriptsize%
		\includegraphics[scale=0.085]{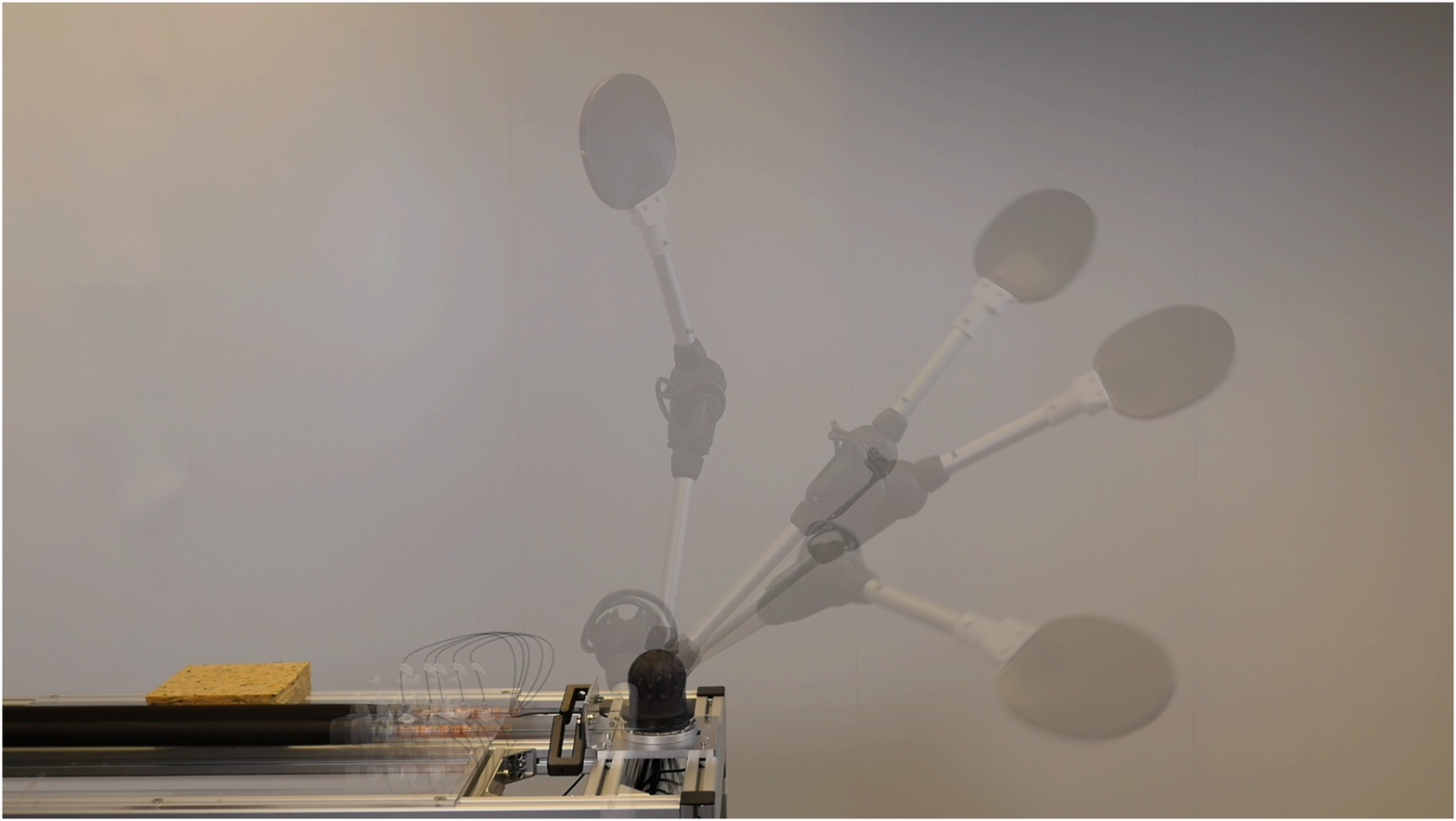}
	}
	\subfloat[time~$=\SI{4}{\second}\ldots\SI{5}{\second}$]{
		\centering\scriptsize%
		\includegraphics[scale=0.085]{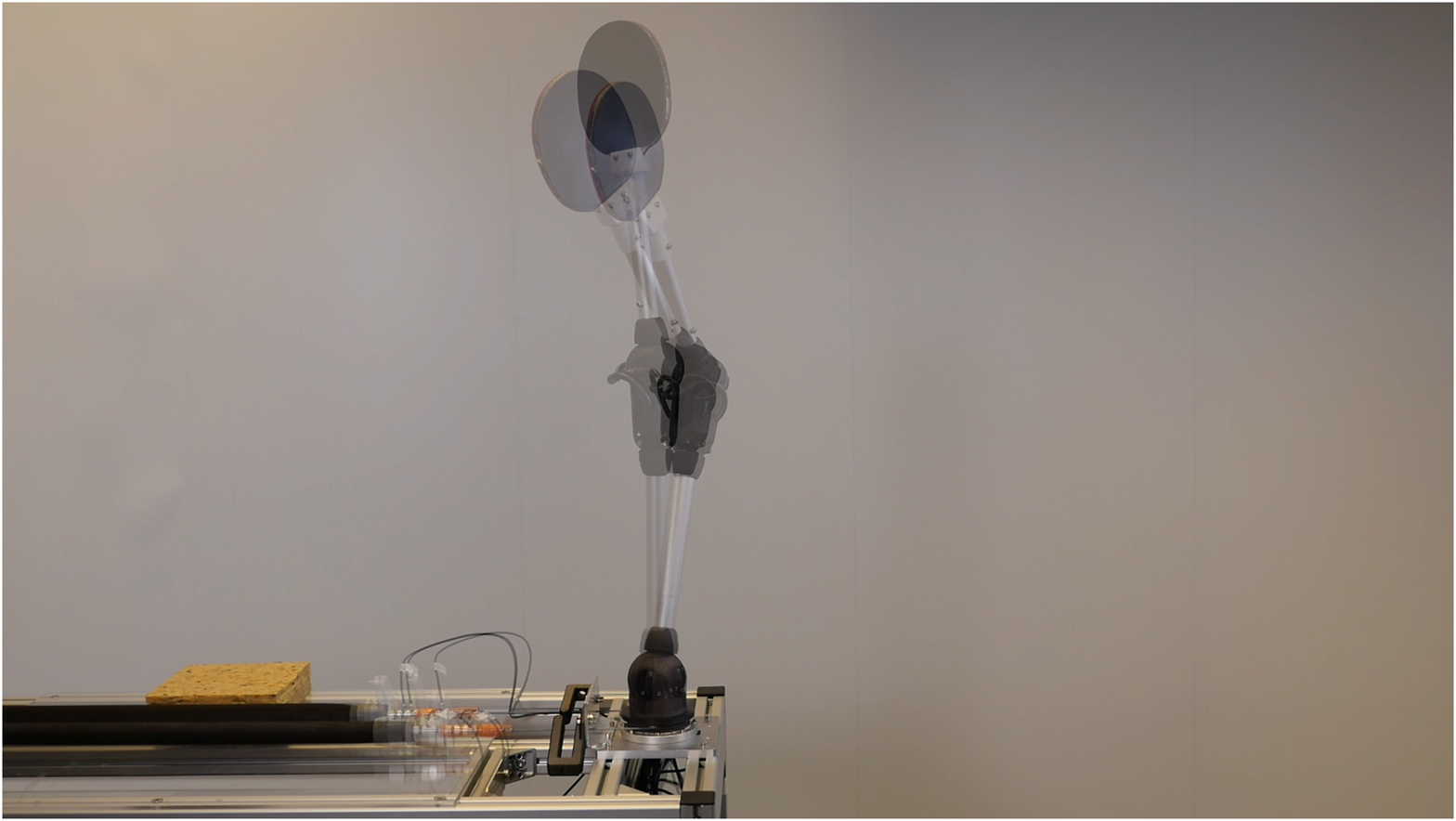}
	}
	\caption{
		Images extracted from a video showing the trajectory tracked in \fig{fig:botrack}. 
		The images represent distinctive phases of the learned motion: (a) zero position, (b) move to start position, (c) hitting motion and (d) move back to zero position. 
	}
	\label{fig:video}
\end{figure*}
\setlength{\figwidth }{0.44\columnwidth} 
\setlength{\figheight }{.618\figwidth}  
\begin{figure*}
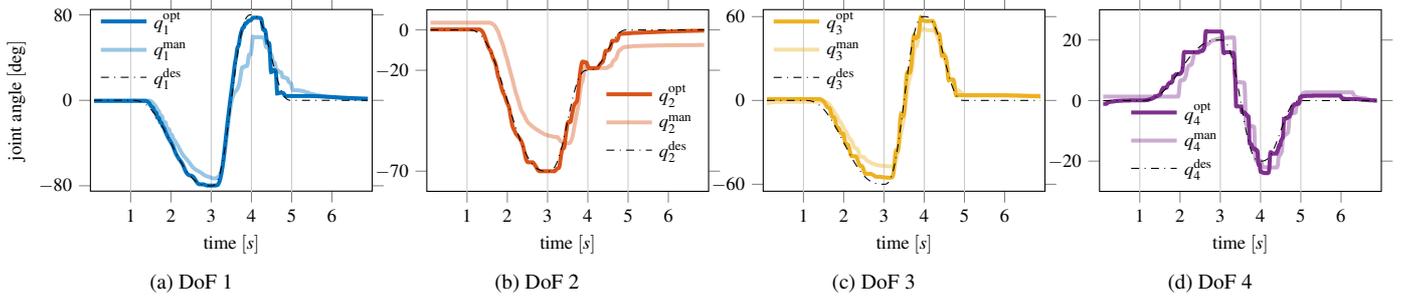

	\centering
	\textbf{Tracking performance on fast trajectory after tuning with Bayesian optimization}\par\medskip
	\vspace{-.2cm}
	\hspace{-.2cm}
	\subfloat[DoF~1]{
		\centering\scriptsize%
		\input{overall1}
	}
	\hspace{-.6cm}
	\subfloat[DoF~2]{
		\centering\scriptsize%
		\input{overall2}
	}
	\hspace{-.6cm}
	\subfloat[DoF~3]{
		\centering\scriptsize%
		\input{overall3}
	}
	\hspace{-.6cm}
	\subfloat[DoF~4]{
		\centering\scriptsize%
		\input{overall4}
	}
	\caption{Tracking performance for all degrees of freedom after optimization using Bayesian optimization~(indicated by the subscript 'opt') and after manual tuning by an expert on the system~(indicated by the subscript 'man'). 
		The trajectory tracked resembles an example of a fast hitting motion between $t=[\SI{3}{\second}, \SI{4}{\second}]$.
		It is composed of a slow motion towards the start position~(\SI{1}{s}$\ldots$\SI{3}{s}) followed by a fast hitting motion~(\SI{3}{s}$\ldots$\SI{4}{s}) and a motion back to the zero position~(\SI{4}{s}$\ldots$\SI{5}{s})~(see \fig{fig:video}).
		The manual parameters are used to track the currently not optimized DoFs in \alg{alg:bo} as $\tman{}$.  
		It is apparent that tracking quality for DoFs three and four is more impaired as for DoFs one and two due to higher friction as the cables are guided through the arm.
		Additionally, the PID with feedforward compensation assume a linear system, hence, BO optimizes towards higher gains as the system is heavily nonlinear.}
	\label{fig:botrack}
\end{figure*}
The aim of this experiment is to highlight the low controlling demands of the arm by showing that adequate tracking performance is possible on slow trajectories using linear controllers only. 
Therefore, we track all four DoFs simultaneously for two kinds of reference signals as can be seen in \fig{fig:tracking}. 
In \fig{fig:tracking}c and \fig{fig:tracking}d a truncated triangular signal was tracked for \SI{10}{\second} and \SI{20}{\second}, respectively. 
The controller from \eq{eq:compfb} has been used and the co-contraction parameter from \eq{eq:dpcorr} is $p_0=0$.
All graphs show that for rapidly changing references, tracking becomes inaccurate.
This deficiency is caused by the PIDs assuming a linear system while PAMs are inherently nonlinear. 
Additionally, for abrupt corrections, in the first moments the change of pressure in the PAM does not affect the joint angle in case the PAMs are not co-contracted enough~\cite{buchler_control_2018}.
For severe cases, this forbearance is followed by a too strong correction as can be seen for DoF two in \fig{fig:tracking}a and c for the middle part of the graph. 
This DoF drives the most mass and hence is harder to control precisely compared to the other DoFs.
Sub-figures \ref{fig:tracking}b and d show tracked sinusoidal references with 0.05 and \SI{0.1}{\hertz}.
Here the same issues occur for rapid changes of the reference.
However, for smooth changes, the reference can be followed with some small delay with all DoF.
\subsection{Generation of Ballistic Movements}
\label{ssec:maxvelacc}
High accelerations are necessary to reach high velocities on a short distance to enable a versatile bouquet of possible trajectories and fast reactions. 
Our system can generate high velocities and accelerations due to the strength of the PAMs used while being robust as a result of the antagonistic muscle configuration.
This property is critical for exploration of fast hitting motions using learning control methods.
\begin{table}[b]
	\caption{Root Mean squared error~(RMSE) comparison of tracking objectives for manually tuned parameters~(MANUAL, $\tman{}$) from \fig{fig:botrack} and tracking with parameters found with our Bayesian optimization approach~(\alg{alg:bo}) from \fig{fig:botrack}~(OPTIMIZED, $\topt{}$).}
	\label{tab:perfcomp}
	\vspace{-.5cm}
	\begin{center}
		\begin{tabular}{c|c c}
			\# DoF & RMSE manual   & RMSE optimized  \\
			\hline
			1 & 12.08 & 3.29\\
			2 & 11.06 & 2.00\\
			3 & 7.22 & 3.74\\
			4 &  3.47 & 2.11
		\end{tabular}
	\end{center}
\end{table}
In this experiment, we show that the system is capable of sustaining the fastest possible motion that can be generated with our system using the rotational DoFs two and three.
The respective minimum pressure was set to one of the PAMs of each muscle pair while the maximum pressure was assigned to the antagonist. 
The subsequent switching from maximum to minimum and vice versa generated a fast trajectory at the end-effector as can be seen in \fig{fig:highvelacc}a.
Note that this step set signal generates the fastest movement at the end-effector that a closed loop controller could have determined without instabilities. 
We did not find any other set signal that moved the arm that close to its joint limits and generated such high peak velocities and accelerations.
The task space $\mathbf{x}=[x_1,x_2,x_3]^T$ has been determined from the joint space coordinates $\mathbf{q}=[q_1,q_2,q_3,q_4]^T$ for each data-point using the forward kinematics equations $\mathbf{x}=T^q_x( \mathbf{q})$.
The forward kinematics equations can be derived from \fig{fig:system}b.
We do not consider the orientation of the end-effector here. 
The resulting velocity and acceleration profiles, depicted in \fig{fig:highvelacc}b and c, show at their respective maxima approximately \SI{12}{\meter\per \second} and \SI{200}{\meter\per \second\squared}.
As a comparison, the fast Barrett Wam arm used for table tennis in~\cite{mulling_biomimetic_2011}, can generate peak velocities of \SI{3}{\meter\per\second} and peak accelerations of \SI{20}{\meter\per\second\squared}. 
The resulting angular velocities in DoF three reaches up to \SI{1400}{\degree\per\second} and angular acceleration of \SI{28000}{\degree\per\second\squared}.
%
\setlength{\figwidth }{0.45\columnwidth} 
\setlength{\figheight }{.618\figwidth}  
\begin{figure*}
	\centering
	\textbf{Similar co-contraction levels appear close to the PF in the objective space}\par\medskip
	\vspace{-.5cm}
	\subfloat[DoF 1]{
		\hspace{-.35cm}
		\centering\scriptsize%
		\input{pareto1}
	}
	\subfloat[DoF 2]{
		\hspace{-0.65cm}
		\centering\scriptsize%
		\input{pareto2}
	}
	\subfloat[DoF 3]{
		\hspace{-.65cm}
		\centering\scriptsize%
		\input{pareto3}
	}
	\subfloat[DoF 4]{
		\hspace{-.45cm}
		\centering\scriptsize%
		\input{pareto4}
	}
	\caption{
		Objective space that spreads the velocity $\Lv$ and position objective $\Lp$ from \eq{eq:lv} and \eq{eq:lp}for all four DoFs where each point represents one tracking instance of the trajectory from \fig{fig:botrack}. 
		The color indicates the value of co-contraction parameter $p_0$. 
		Note that the figures are zoomed in to illustrate the estimated Pareto front~(PF), hence, differently colored points lie outside this area.
		It is apparent that points with similar $p_0$ appear close to the estimated PF instead of being diverse. 
		Substantially different colors are almost not present~(except of for DoF three) as they lie outside the zoomed area.
		The dominant co-contraction range close to the PF is $p_{0,1}=0.9\ldots 1$ for DoF one~(a), $p_{0,2}=0.5\ldots 0.7$ for DoF two~(b), $p_{0,3}=0.2\ldots 0.5$ for DoF three~(c) and $p_{0,4}=0.6\ldots 0.9$ for DoF four~(d).
	}
	\label{fig:pareto}
\end{figure*}
\subsection{Bayesian Optimization of Controller Parameters}
\label{ssec:botrack}
Having demonstrated that the robot arm can be controlled using simple PIDs for slow trajectories and that the system sustains fast hitting motions, the natural next goal is to learn to control fast trajectories~(see \fig{fig:video}). 
Tuning control parameters at higher speeds leads to potentially dangerous configurations on traditional motor-driven systems that we can partially avoid using antagonistic actuation as discussed in \sec{ssec:pid}. 
We tune seven parameters
\begin{align}
\tta{i}=[k^\text{P}_i,k^\text{D}_i,k^\text{I}_i,k^\text{pos}_i,k^\text{vel}_i,k^\text{acc}_i,p_{0,i}]
\label{eq:BOpara}
\end{align} 
for each of the four DoFs $i=1\dots4$~(28 parameters in total) where the additional feedforward components from \eq{eq:compff} improve control on fast trajectories.

In addition to the parameter of the feedback and feedforward components, we also optimize the co-contraction parameter $p_0$ from \eq{eq:dpcorr}. 
Too much co-contraction increases the friction within the tendon-driven system whereas too low $p_0$ complicates control as the tendons are not stretched for some configurations.
Hence, we aim at answering the question of whether the stiffness in a joint - using the co-contraction $p_0$ as a proxy - influences the tracking significantly. 
\setlength{\figwidth }{0.9\columnwidth} 
\setlength{\figheight }{.618\figwidth}  
\begin{figure}[t]
	\centering
	\textbf{Minimum overall objective trace}\par\medskip
	\scriptsize
	\input{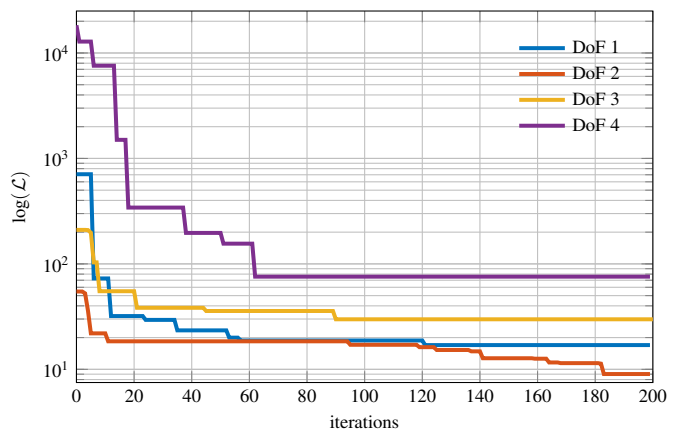}
	\caption{
		Minimum overall objective trace over 200 iterations for each DoF.
		All DoFs start at different initial values as the parameter for the first iteration are drawn randomly. 
		After approximately 60 iterations the performance has improved by at least one order of magnitude. 
		The main contributer is that our approach uses additionally potentially dangerous data points which would break traditional motor-driven systems. 
	}
	\label{fig:mintrace}
\end{figure}
\noindent

Optimizing 28 parameters in total renders manual tuning impossible.
\fig{fig:botrack} depicts the tracking performance that is achieved after two hours of manual tuning~($q^\text{man}_i$) by an expert on the system.
Especially troublesome is the interdependence of the DoFs.
Tuning one DoF to a sufficient performance often impairs tracking of the other DoFs. 
Hence, we incorporate the BO tuning scheme from \alg{alg:bo}~(described in~\sec{ssec:autotune}) to tune one DoF at a time and apply either manual parameters $\tman{i}$ or the optimized parameters $\topt{i}$ in case they were already found to the other DoFs $i$.
We use the position $\Lp$ and velocity losses $\Lv$ from \eq{eq:lp} and \eq{eq:lv} respectively as performance measure.
Although we also employ the action loss $\La$~(\eq{eq:la}), it merely acts as a regularizer to allow all curvatures of the action signal within the allowed range $[-1,1]$ while penalizing exceeding these limits. 
\fig{fig:botrack} illustrates the tracking performance found~($q^\text{opt}_i$) by applying \alg{alg:bo} and \tab{tab:perfcomp} compares the mean squared error between the corresponding tracking performances.
The algorithm converges quickly after 200 iterations for each DoF as can be seen in \fig{fig:mintrace}.
To the best of our knowledge, this is the fastest tracked trajectory with a four DoF PAM-based robot~(see \fig{fig:video} for an illustration of the trajectory).
We think that the main advantage gained is that our BO tuning procedure additionally learns from data points that would break traditional robotic systems. 
As a result, our approach pioneers the direct application of Bayesian optimization on a real system on a task that potentially breaks the system. 

The amount of co-contraction $p_0$ is an essential property of PAM-based systems.
An infinite set of pressures in one PAM pair leads to the same joint angle; the co-contraction discriminates this infinite set.
Also, it correlates with the stiffness in the DoF. 
An interesting question is whether the right choice of the co-contraction improves the control performance.
To answer this question, we study the connection between loss and co-contraction.
From all the points for each DoF, we calculated the estimated Pareto front~(PF) using \eq{eq:pareto} and colored the data depending on the value of the co-contraction parameter $p_0$. 
\fig{fig:pareto} shows the objective space for each DoF where one point represents one tracking instance. 
The figure spans the position $\Lp$ as well as the velocity objective $\Lv$ from \eq{eq:lp} and \eq{eq:lv}.
We left out the regularizing action objective $\La$.
Not all data points are visible as the figures are zoomed to illustrate the PF more clearly. 
It is apparent that close to the PF the values of $p_0$ are similar.
In particular, for DoF one, two and four significantly different colors do not appear in the figure at all as they lie outside the zoomed-in area.
The absence of a whole range of co-contraction levels is a strong hint that by choosing $p_0$ conveniently, a linear controller can better control our generally nonlinear system.
	
	\section{Conclusion}
	\label{sec:conclusion}
Generating high accelerations with robotic hardware as seen in human arm flick motions, while maintaining safely during the learning process, are desirable properties for modern robots~\cite{buchler_lightweight_2016}.
In this paper, we built a lightweight arm actuated by PAMs to try to address this issue. 
Our robot avoids key problems of previous PAM-driven arms, such as unnecessary directional change of cables as well as PAMs that are bent around or touch the structure or other PAMs.
We experimentally show that our system eases control by tracking a slow trajectory sufficiently well while incorporating only simple and manually tuned linear PID controllers.
Execution of ballistic movements illustrate that our arm surpasses the peak task space velocity and acceleration of a Barrett Wam arm by a factor of 4x and 10x respectively.
Experiments also show that our system can sustain the large forces generate by high accelerations and antagonistic actuation.
This property allows machine learning algorithms to explore in fast regimes without taking safety into account. 
To demonstrate this ability, we automatically tuned PID controllers with additional feedforward components using BO to learn to track a fast trajectory.
Predefined pressure and parameter limits were sufficient to avoid damaging the robot itself and prevent unbounded instabilities during training. 
We did not find any other reported publication that tracked faster trajectories with a comparable PAM-driven system. 
Data points collected during training indicate that trials close to the estimated PF own similar co-contraction levels. 
In this manner, we empirically illustrate that the choice of the co-contraction level has a significant influence on the performance at trajectory tracking tasks. 

Interesting future directions are to extend the system to six DoFs to allow to reach arbitrary positions and orientations with the end-effector. 
We are also curious to learn low-level control~(in muscle space), for instance using reinforcement learning, to perform more complex tasks that involve fast motions. 
Another interesting direction is to extend the research on how to make use of the overactuation derived from the co-contraction such as in~\cite{buchler_control_2018}.
Lastly, it is intriguing to understand to what extent pretrained policies in simulation can be used directly on the real system for high-acceleration tasks.

	
	%
	%


	 \appendix
	
	  \section{Software Framework}\label{ssec:software}
	 The complete system comprises eight pressure sensors and proportional valves as well as four incremental angular encoders to govern and sense the movement.
	 Each DoF is actuated by two antagonistically aligned PAMs. 
	 The contraction ratio as well as the pulling force is influenced by the air pressure within each PAM. 
	 Thus, a low level controller regulates the pressure within each PAM using Festo proportional valves as can be seen in \fig{fig:cascade}.
	 As a result, the control algorithm that regulates the movement works on top of these and sends desired pressures $\pdes{}$ to control the joint angles $\mathbf{q}$.
	 A National Instruments PCIe 7842R FPGA card has been used to take over low level tasks such as extraction of the angular values from the A und B digital signals given by the encoder or regulating the pressure within each PAM. 
	 The FPGA was programmed in Labview.
	 To assure fast implementation, we used the FPGA C/C++ API interface to generate a bitfile along with header files which can be incorporated in any C++ project. 
	 Thus, the control algorithm can be implemented in C++ on top of the basic functionalities supplied by the FPGA.
	 The sensor values are read at \SI{100}{\kilo\hertz} and new desired pressure values are adjusted at \SI{100}{\hertz}. 
	 \fig{fig:system}a shows the pressure response to a step in desired value from minimum~(\SI{0}{\bar}) to maximum air pressure~(\SI{3}{\bar}). 
	 The resulting pressure regulation reaches the desired value within a maximum of \SI{0.25}{\second}.
	\section*{References}
	\bibliography{refsnourl}
	
\end{document}